\documentclass[sn-apa, iicol]{sn-jnl}% APA Reference Style 
\usepackage{graphicx}%
\usepackage{multirow}%
\usepackage{amsmath,amssymb,amsfonts}%
\usepackage{amsthm}%
\usepackage{mathrsfs}%
\usepackage[title]{appendix}%
\usepackage{xcolor}%
\usepackage{textcomp}%
\usepackage{manyfoot}%
\usepackage{booktabs}%
\usepackage{algorithm}%
\usepackage{algorithmicx}%
\usepackage{algpseudocode}%
\usepackage{listings}%
\usepackage{natbib}
\usepackage{caption, subcaption}
\usepackage{arydshln}
\usepackage{pifont}
\usepackage{acronym}
\acrodef{OMR}{Optical Music Recognition}
\acrodef{OCR}{Optical Character Recognition}
\acrodef{HTR}{Handwritten Text Recognition}
\acrodef{CTC}{Connectionist Temporal Classification}
\acrodef{MEI}{Music Encoding Initiative}
\acrodef{CNN}{Convolutional Neural Network}
\acrodef{CER}{Character Error Rate}
\acrodef{SER}{Symbol Error Rate}
\acrodef{LER}{Line Error Rate}
\acrodef{SMT}{Sheet Music Transformer}
\acrodef{PE}{Positional Encoding}
\acrodef{DAN}{Document Attention Network}

\newcommand{\krn}{\textsc{kern}}
\newcommand{\bkrn}{\textsc{bekern}}
\newcommand{\ekrn}{\textsc{ekern}}
\newcommand{\review}[1]{\textcolor{black}{#1}}

\newcommand{\GrandStaff}{\textsc{GrandStaff}}
\newcommand{\FPSMTCNN}{$\text{SMT}_{\text{CNN}}$}
\newcommand{\FPSMTNXT}{$\text{SMT}_{\text{NeXt}}$}
\newcommand{\FPSMTNXTff}{$\text{SMT}_{\text{NeXt}}^{f}$}

\newcommand{\xmark}{\ding{55}}%
\newcommand{\cmark}{\ding{51}}%

\theoremstyle{thmstyleone}%
\theoremstyle{thmstyletwo}%

\theoremstyle{thmstylethree}%

\begin{document}

\title[End-to-End Full-Page Optical Music Recognition for Pianoform Sheet Music]{End-to-End Full-Page Optical Music Recognition for Pianoform Sheet Music}


\author*[1]{\fnm{Antonio} \sur{Ríos-Vila}}\email{arios@dlsi.ua.es}

\author[1]{\fnm{Jorge} \sur{Calvo-Zaragoza}}\email{jcalvo@dlsi.ua.es}

\author[1,2]{\fnm{David} \sur{Rizo}}\email{drizo@dlsi.ua.es}

\author[3]{\fnm{Thierry} \sur{Paquet}}\email{thierry.paquet@litislab.eu}

\affil*[1]{\orgdiv{Pattern Recognition and Artificial Intelligence Group}, \orgname{University of Alicante}, \orgaddress{\country{Spain}}}

\affil[2]{\orgdiv{Instituto Superior de Enseñanzas Artísticas de la Comunidad Valenciana}, \orgaddress{\country{Spain}}}

\affil[3]{\orgdiv{LITIS Laboratory - EA 4108}, \orgname{University of Rouen}, \orgaddress{\country{France}}}


\abstract{Optical Music Recognition (OMR) has made significant progress since its inception, with various approaches now capable of accurately transcribing music scores into digital formats. Despite these advancements, most so-called \emph{end-to-end} OMR approaches still rely on multi-stage processing pipelines for transcribing full-page score images, \review{which entails challenges such as the need for dedicated layout analysis and specific annotated data, thereby limiting the general applicability of such methods}. In this paper, we present the \review{first truly end-to-end approach for page-level OMR in complex layouts}. Our system, which combines convolutional layers with autoregressive Transformers, processes an entire music score page and outputs a complete transcription in a music encoding format. This is made possible by both the architecture and the training procedure, which utilizes curriculum learning through incremental synthetic data generation. \review{We evaluate the proposed system using pianoform corpora, which is one of the most complex sources in the OMR literature}. This evaluation is conducted first in a controlled scenario with synthetic data, and subsequently against two real-world corpora of varying conditions. Our approach is compared with leading commercial OMR software. The results demonstrate that our system not only successfully transcribes full-page music scores but also outperforms the commercial tool in both zero-shot settings and after fine-tuning with the target domain, representing a significant contribution to the field of OMR.}

\keywords{Optical Music Recognition, Sheet Music Transformer, Full-Page Transcription, Document Image Analysis}

\maketitle

\section{Introduction}\label{sec:intro}
Music is a vital part of our cultural heritage, offering insights into the social, cultural, and artistic trends of various historical periods. For this reason, many music documents have been carefully preserved over the centuries. To safeguard this cultural heritage and democratize access to it, significant efforts have been made toward digitizing these documents and storing them as digital copies (images). However, this format does not allow for the full exploitation of the number of applications in computational musicology, which require the music-notation content to be transcribed into a music encoding format, such as MusicXML---among others. Due to the high cost of manual transcription, producing large historical archives by hand becomes impractical, necessitating the development of automatic processes to achieve this goal. The field of \ac{OMR} is dedicated to the computational reading of music scores~\citep{Calvo-Zaragoza:ACM:2020}. An OMR system typically takes an input music score and seeks to produce a symbolic transcription, which is then encoded in a standard digital format that represents its music-notation content.

The way \ac{OMR} has been approached has evolved throughout its history, from multi-stage statistical learning pipelines~\citep{Bainbridge:CAH:2001, Rebelo:IJMIR:2012} to \textit{end-to-end} deep learning-based solutions. Currently, there are two main methodological perspectives for transcribing full-page music scores. The first involves note retrieval approaches, which are primarily based on object detection. These approaches mainly extract notation primitives from full-page score images~\citep{Pacha:ICDAR:2017,Song:SIPNC:2022,DeVega:CEC:2022,Hartelt:Algorithms:2022,Yesilkanat:DAS:2024}.
The second are the holistic approaches, where transcription is treated as a sequence generation problem. In this family, a sequential representation of the score is expected to be directly produced from the input image~\citep{Calvo-Zaragoza:ISMIR:2018,Calvo-Zaragoza:PRL:2019,Alfaro-Contreras:PRL:2022,Torras:ISMIR:2022}.

Despite the significant progress in \ac{OMR}, these approaches are still dependent on pipeline-based workflows when processing full-page scores. Note retrieval methods fail to directly reproduce the complete content of the score, as they output a collection of music primitives that require further assembly and encoding~\citep{Rossant:JASP:2006,Schuiling:CMR:2020,Baro:ICFHR:2022,Penarrubia:ISMIR:2023}. Furthermore, holistic approaches typically assume that the input is a staff-level image, relying on a prior layout analysis step to extract individual music staves for separate processing and subsequent post-concatenation~\citep{Campos:ICFHR:2016,Castellanos:ISMIR:2020,Castellanos:ESWA:2022}. These methods sometimes do not even produce a standard digital music encoding as the final representation of the score but rather an approximation of it, necessitating additional steps for encoding~\citep{RiosVila:APPSCI:2021,RiosVila:ICDAR:2021}. \review{Recently, there have been some attempts to perform full-page end-to-end transcription through single models, although these were tested in very simple scenarios with multiple constrains~\citep{RiosVila:ISMIR:2022}.} Thus, although many OMR methods are presented as end-to-end in the literature, they largely fall short of this claim. 

\review{In this paper, we present the first end-to-end full-page OMR transcription model, capable of producing digital music scores from high-complexity images}. \review{This solution is based on autoregressive Transformers and curriculum learning~\citep{Vaswani:NIPS:2017}. While our architecture builds upon the principles of the \ac{DAN} model~\citep{Coquenet:TPAMI:2023}, originally introduced for handwritten text recognition, our contribution lies in successfully adapting and extending this approach to the domain of full-page \ac{OMR}. Unlike prior work, our system handles the graphical and symbolic complexity of music notation—including hierarchical structures, dense symbolic content, and long-range dependencies—without relying on intermediate processing stages. To the best of our knowledge, this represents the first demonstration of a model operating effectively in this highly structured and underexplored domain.} We test this approach on the pianoform transcription task, which is the most complex musical texture---both in graphical and notational terms---and the one that draws the most interest \citep{Calvo-Zaragoza:ACM:2020}. The performance of the model is assessed through two experimental scenarios: a controlled scenario using synthetic data to determine the best model configuration, and a real-world scenario using two publicly available historic pianoform score datasets. \review{The second scenario includes a thorough comparison with well-established commercial and open-source OMR tools, as well as state-of-the-art pipelines}. Experiments demonstrate that our model is not only capable of performing full-page pianoform transcription but also \review{outperforms current state of the art and reaches competitive results against top-leading commercial software, setting a new milestone in the OMR field}.

The remainder of the paper is structured as follows: Section \ref{sec:methodology} presents the neural network architecture of the proposal. Section \ref{sec:encoding} discusses and justifies the standard output encoding of the model. In Section \ref{sec:cl} we describe the curriculum process designed to train the \ac{SMT} to be able to transcribe full-page sources. Then, Section \ref{sec:experiments} describes the setup designed to assess the performance of the \ac{SMT}. The scenarios and the results analysis are given in Section \ref{sec:results}. Finally, a summary of the work and final conclusions are given in Section \ref{sec:conclusions}. 

\section{Sheet Music Transformer}
\label{sec:methodology}
\begin{figure*}[h]
\centering
    \includegraphics[width=\textwidth]{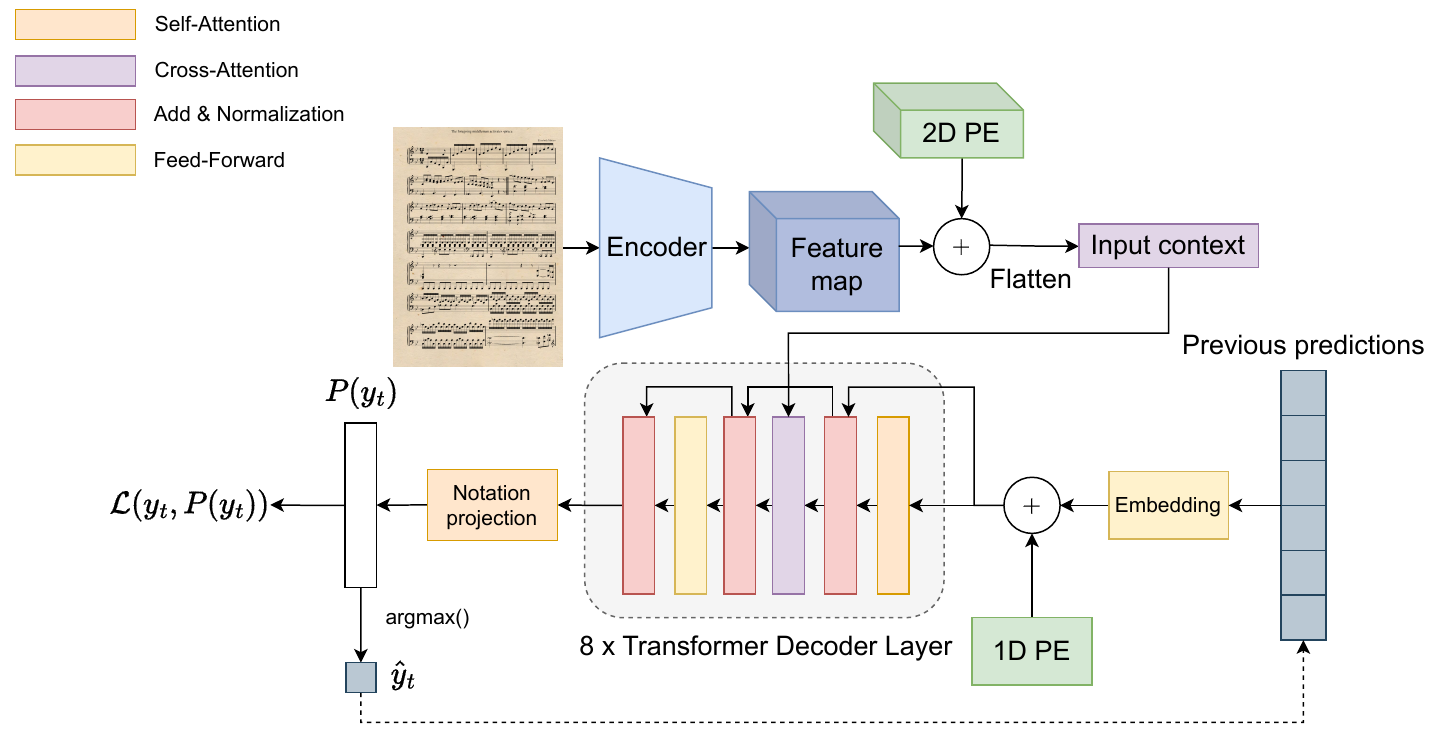}
    \caption{Graphic scheme of the \ac{SMT} for end-to-end full-page Optical Music Recognition.}
    \label{fig:FP-SMT}
\end{figure*}

The approach presented in this paper is built upon the \acf{SMT}, an autoregressive end-to-end neural network---\review{inspired by the \ac{DAN} architecture}~\citep{Coquenet:TPAMI:2023}---that generates a transcription of the music score image provided as input. The model consists of two fundamental components: an encoder and a decoder. The encoder is a feature extractor that processes the image $x$ and produces a meaningful representation $x_e^{\prime}$. The decoder is an autoregressive language model that predicts at each time step the \review{next} symbol, based on the maximum a posteriori decision rule. In this respect the decision is conditioned on the prefix (previously generated symbols) and the representation from the encoder. This process is formalized as: 

\begin{equation}
\label{eq:process_comp}
    \hat{y}_t = \arg\max_{y_t \in \Sigma} P(y_t \mid x_e^{\prime}, \left(\hat{y}_0, \hat{y}_1, \ldots, \hat{y}_{t-1}\right))
\end{equation}

where $\Sigma$ represents the music-notation vocabulary of a music encoding language, and $t$ is the current timestep. A graphical representation of the \ac{SMT} is shown in Fig. \ref{fig:FP-SMT}.

\subsection{Encoder}
Let $x \in \mathbb{R}^{c \times h \times w}$ represent the input image of the \ac{SMT}, where $h$ and $w$ denote its height and width in pixels, respectively, and $c$ denotes the number of channels. To learn an appropriate representation that can be used by the Transformer decoder, we employ a Convolutional Neural Network (CNN), as considered in recent literature~\citep{Singh:ICDAR:2021, Coquenet:TPAMI:2023}. This module, therefore, defines a function that outputs $c_e$ two-dimensional feature maps, denoted as $x^{\prime}_e \in \mathbb{R}^{h_e \times w_e \times c_e}$. Note that $h_e$ and $w_e$ correspond to the size of the learned feature maps, where $h_e = \frac{h}{r_h}$ and $w_e = \frac{w}{r_w}$, with $r_h$ and $r_w$ representing the downscaling factors produced by the pooling operations and strides within the encoder, while $c_e$ is the dimension of the feature space of the encoder.

\subsection{Decoder}
\label{sec:decoder}
The second module of the \ac{SMT} is a Transformer decoder~\citep{Vaswani:NIPS:2017}, as it represents the current state of the art for tasks involving conditional sequence generation of varying lengths.

The decoder consists of a language model that, at each timestep $t$, computes the conditioned probability distribution $p_t$ defined at Equation \ref{eq:process_comp}.
%outputs a probability distribution $p_t \in \mathbb{R}^{\left|\Sigma\right|}$ over the $\Sigma$ vocabulary symbols. This is conditioned on the input feature vector produced by the encoder, $x_e^{\prime}$, along with the previously predicted tokens, $\left(\hat{y}_0, \ldots, \hat{y}_{t-1}\right)$. 
The predicted token $\hat{y}_t$ is the one that maximizes this probability score. The process starts with a special start-of-sequence symbol, $\hat{y}_0 = \text{\textless \textit{sos}\textgreater}$, and continues until an end-of-sequence token is predicted ($\hat{y}_{|\mathbf{\hat{y}}|} = \text{\textless \textit{eos}\textgreater}$).

We emphasize that the encoder outputs a two-dimensional feature map to serve as context for the decoder. However, since the decoder is a sequence generator, it operates in just one dimension. To connect these two modules, the two-dimensional feature maps are converted into a one-dimensional format suitable for the decoder. A straightforward approach is to flatten the structure across height and width, resulting in sequences of length $h_e \times w_e$.

Another important aspect is that the Transformer decoder includes a positional encoding (PE) mechanism to model its otherwise order-agnostic operation. This mechanism adds a position vector to each input element, determined by its location in the sequence. While adding a one-dimensional PE to the unfolded feature maps could be considered, it would cause a loss of spatial information when dealing with polyphonic scores where multiple voices are played simultaneously. To ensure the model retains awareness of the full spatial dimensions of the image, we incorporate a two-dimensional PE within the feature maps before flattening them into a one-dimensional sequence. The 2D PE is based on sine and cosine functions, similar to the original 1D PE, where the first half of the feature dimensions---i.e., $[0, \frac{c_e}{2})$---is responsible for horizontal positions, and the second half---i.e., $[\frac{c_e}{2}, c_e)$---is used for vertical positions, as implemented in other works~\citep{Singh:ICDAR:2021,Coquenet:TPAMI:2023}. The 2D PE is defined as:

%\frac{pos}{1000^{2i / c_e})

\begin{align}
\begin{split}
\label{eq:PE2D}
    \text{PE}_{\text{2D}}\left(\text{pos}_t, 2i\right) & =\sin \left(\frac{\text{pos}_t}{10000^{2i / c_e}}\right) \\
    \text{PE}_{\text{2D}}\left(\text{pos}_t, 2i + 1\right) & =\cos \left(\frac{\text{pos}_t}{10000^{2i / c_e}}\right) \\
    \text{PE}_{\text{2D}}\left(\text{pos}_f, c_e/2 + 2i\right) & = \sin \left(\frac{\text{pos}_f}{10000^{2i / c_e}}\right) \\
    \text{PE}_{\text{2D}}\left(\text{pos}_f, c_e/2 + 2i + 1\right) & = \cos \left(\frac{\text{pos}_f}{10000^{2i / c_e}}\right)
\end{split}
\end{align}

where $\text{pos}_t$ and $\text{pos}_f$ specify the horizontal (width) and vertical (height) positions, respectively, and $i \in\left[0, \frac{c_e}{4} \right)$ denotes the feature dimension of the output map.

\section{Output Encoding}
\label{sec:encoding}
There are numerous options when choosing a digital music encoding format, being MusicXML~\citep{Good:XML:2001} the most popular one. This format is known for its comprehensive representation of music score components and metadata through an XML-based markup language. Despite being highly versatile, it is not well-suited for deep learning-based solutions. The reason is that it is too verbose and typically require conversions and adaptations into intermediate languages, which are often lossy.

Based on recent literature~\citep{RiosVila:DLFM:2020,RiosVila:APPSCI:2021,RiosVila:IJDARICDAR:2023,RiosVila:IJDARICDAR:2023}, in this work we adopt the text-based Humdrum Kern encoding format~\citep{Huron:BMIDI:1997}, hereinafter referred to as \krn{}. This encoding stands out in computational music analysis due to its widespread use, simplicity, and straightforward parsing capabilities, making it exceptionally well-suited for end-to-end \ac{OMR} applications. Additionally, this format is compatible with many music software packages~\citep{Pugin:ISMIR:2014} and can be converted into other digital music representations, such as MusicXML and others.

\subsection{The Humdrum Kern Format}
\begin{figure}
    \centering
    \includegraphics[width=\columnwidth]{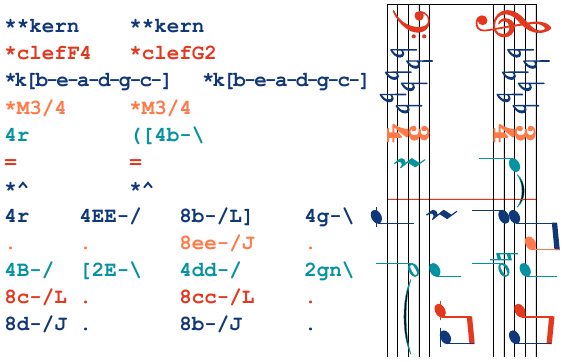}
    \caption{Example of a \krn{} score (left) aligned by reading order with its rendered music document (right).}
    \label{fig:kernsample}
\end{figure}

A \krn{} file is essentially a sequence of text lines.\footnote{For a comprehensive explanation of the syntax, the reader is referred to the official documentation (\url{https://www.humdrum.org/rep/kern/}).} Each line is, in turn, a sequence of columns or \emph{spines} that are separated by a \textit{tab} character. Each column contains an instruction, such as the creation or termination of spines, or the encoding of musical symbols such as clefs, key signatures, meter, bar lines, notes, or chords (which are notes separated by spaces), to name just a few. When interpreting a \krn{} file, all spines are read simultaneously, providing polyphonic capabilities to the format. In other words, a line in a \krn{} document is read from left to right---interpreting all symbols that appear simultaneously---and then from top to bottom, advancing through the score and time. Figure \ref{fig:kernsample} shows an example of how a \krn{} file is constructed, aligned with its corresponding score.

Since the \ac{SMT} contains an autoregressive Transformer decoder, we can use a complete \krn{} file as ground-truth. We also include additional markup tokens to ensure correct understanding of the file structure. Spine separations are encoded as \textless\textit{t}\textgreater, chords as \textless\textit{s}\textgreater, and timesteps as \textless\textit{b}\textgreater.

\subsection{Humdrum Kern Tokenization}
In this work, we explore three different ways to tokenize a \krn{} file for training the \ac{SMT}.

First, we interpret all the complete \krn{} symbols as tokens. In the example from Fig.~\ref{fig:kernsample}, \texttt{8ee-/J} would be a single token. This tokenization, hereinafter referred to as raw \krn{} tokenization, produces a large vocabulary, which surpasses, based on the information in Table \ref{tab:corpus-features} from the evaluated corpora, more than $20,000$ elements. This is inconvenient for this task, \review{as it may lead to training instability}. Therefore, we explore two alternative methods to address this issue. Additionally, each \krn{} token is composed of several semantic subcomponents, making it worthwhile to consider some other tokenization schemes.

Following the work of \cite{RiosVila:IJDARICDAR:2023}, we also explore the ``Basic Extended Kern'' format, referred to as \bkrn{}. This format significantly reduces the alphabet by decomposing tokens into minimal semantic elements, such as duration or pitch of a note. For example, in Fig.~\ref{fig:kernsample}, the \texttt{8cc-/L]} token is encoded as \texttt{8 c··-·/·L}.\footnote{Note that the character `·' is used here solely as a separator for clarity.}

While the \bkrn{} tokenization is optimal for reducing vocabulary size, it suffers from lower correlation with the graphic information in the image, as the model needs to predict several tokens for a single graphical pattern. This becomes problematic when handling large documents. To address this, we also explore the ``Extended Kern''—\ekrn{}, which splits the tokens by graphical meaning. In this way, music notes are represented by a unique token, while accidentals or other related features are represented independently. Continuing with the example from Fig.~\ref{fig:kernsample}, the \texttt{8cc-/L} token is encoded as \texttt{8cc·-·/·L}.

It is important to note that all three tokenization methods---raw \krn{}, \bkrn{}, and \ekrn{}---are equally complete, in the sense that they encode the same musical information, differing only in how the elements are split and fed to the input of the \ac{SMT}. Their suitability for \ac{OMR} tasks will be determined empirically through fair comparison of their performance in our experiments.

%\section{Full-page transcription}
\section{Training Methodology}
\label{sec:cl}
Training Transformer-based models is known to be a data-demanding and challenging task, primarily due to their low inductive bias~\citep{Delvin:ACACL:2019,Brown:NIPS:2020}. Designing an effective training process to handle end-to-end full-page transcription is crucial for the performance of the \ac{SMT}. Inspired by recent literature in related fields~\citep{Coquenet:TPAMI:2023,Dhiaf2023}, the model is trained in a three-stage process:

\begin{enumerate}
    \item System-level pre-training: The \ac{SMT} is initially trained at the system level, using images that depict a single system. This enables the model to learn the first step in the reading order of pianoform music scores.
    \item Incremental curriculum learning: The single system-transcription \ac{SMT} is then trained to read full-page pianoform scores. This is accomplished through an incremental curriculum learning method, where the model is gradually exposed to synthetic images containing an increasing number of systems.
    \item Curriculum learning-based fine-tuning: A second curriculum learning strategy is employed for fine-tuning the \ac{SMT}, in which the model is trained on a combination of synthetic and real music score pages (gradually increasing the proportion of the latter ones). This allows the model to learn the training corpus while retaining the general knowledge acquired from the synthetic samples.
\end{enumerate}

Below, we present the approach used to generate synthetic pianoform music scores before discussing the different training stages in detail.

\subsection{Synthetic Generator}
\label{subsec:synthgen}
The overall training strategy requires the implementation of a synthetic music score generator. This tool is essential for providing a sufficient variety of samples and for controlling the difficulty of the task during curriculum learning. Despite some previous efforts~\citep{Hernandez:SoftwareX:2023, MartinezSevilla:ICDAR:2023, AlfaroContreras:IJMIR:2023} and \review{ongoing projects~\citep{Mayer:ICDAR:2021},   there remains a lack of synthetic music score generators capable of producing samples suitable for end-to-end \ac{OMR} approaches in Humdrum **kern, particularly in the case of pianoform scores.}

Our generator is initialized with a collection of single-system \krn{} excerpts, which the generator automatically parses. To maintain rhythmic consistency, it categorizes these samples by their beat types. The generation algorithm selects a specified number of systems at random and merges them according to \krn{} syntax rules. The clefs and key signatures are determined by the first selected system, with subsequent excerpts adjusted accordingly. Thanks to the \krn{} syntax’s agnosticism toward these elements, this integration process is seamless. The generated file is then rendered using the Verovio tool~\citep{Pugin:ISMIR:2014}, producing the final music score page. In addition to this core algorithm, the generator includes additional features:

\begin{itemize}
    \item Random texture placement to simulate real documents. The generator selects a texture from a library of 61 samples. To place a texture, the algorithm crops the page size and positions the generated music image on it.\footnote{All the texture images are larger than standard A4 pages.}
    \item Randomly generated title and author names.
    \item Varying spacing between staff lines and note font sizes.
    \item Varying inner score padding on the page.
\end{itemize}

\begin{figure*}
     \centering
     \begin{subfigure}[b]{0.32\textwidth}
         \centering
         \includegraphics[width=\textwidth]{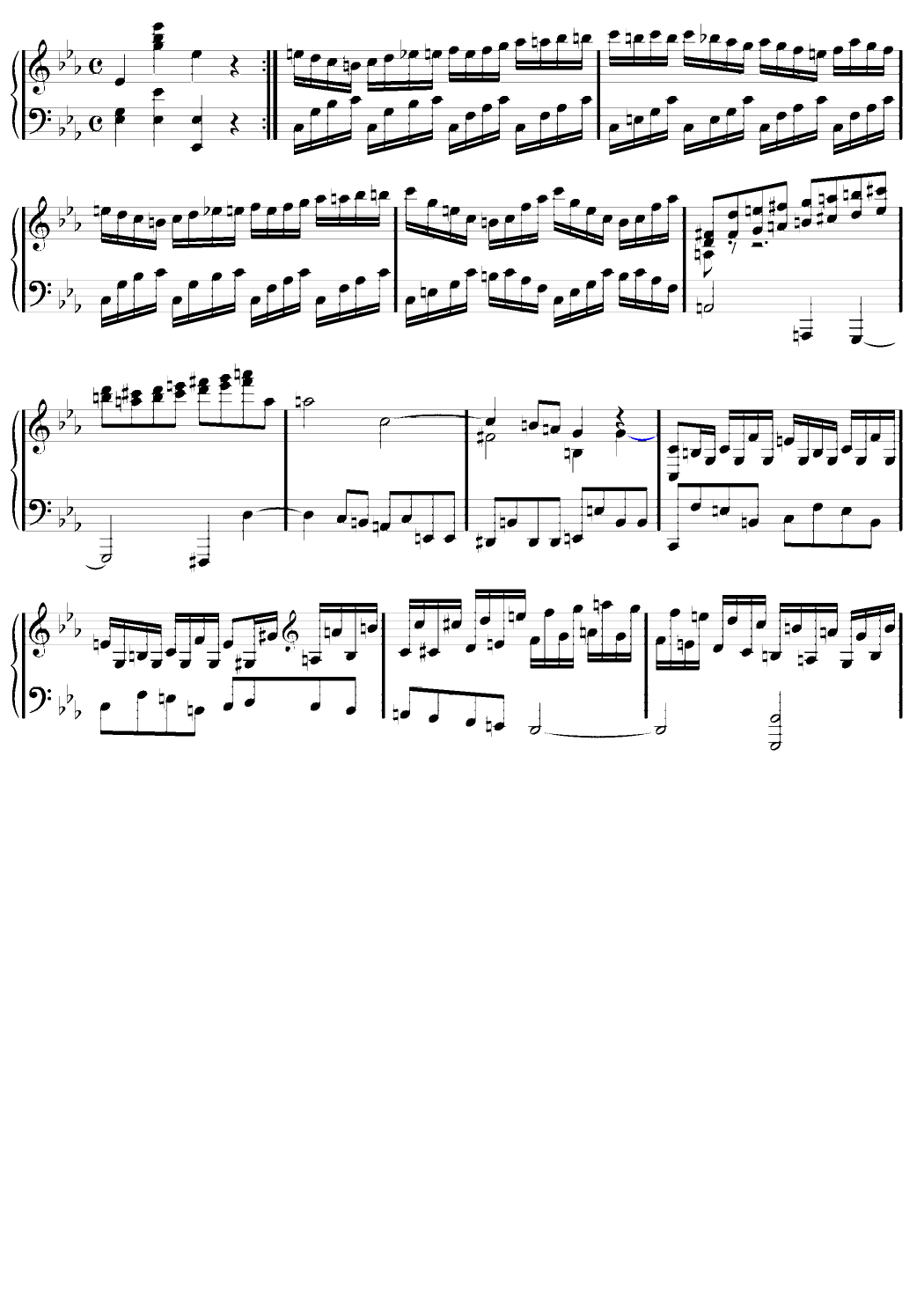}
         \caption{}
         \label{fig:sample_easy}
     \end{subfigure}
     \begin{subfigure}[b]{0.32\textwidth}
         \centering
         \includegraphics[width=\textwidth]{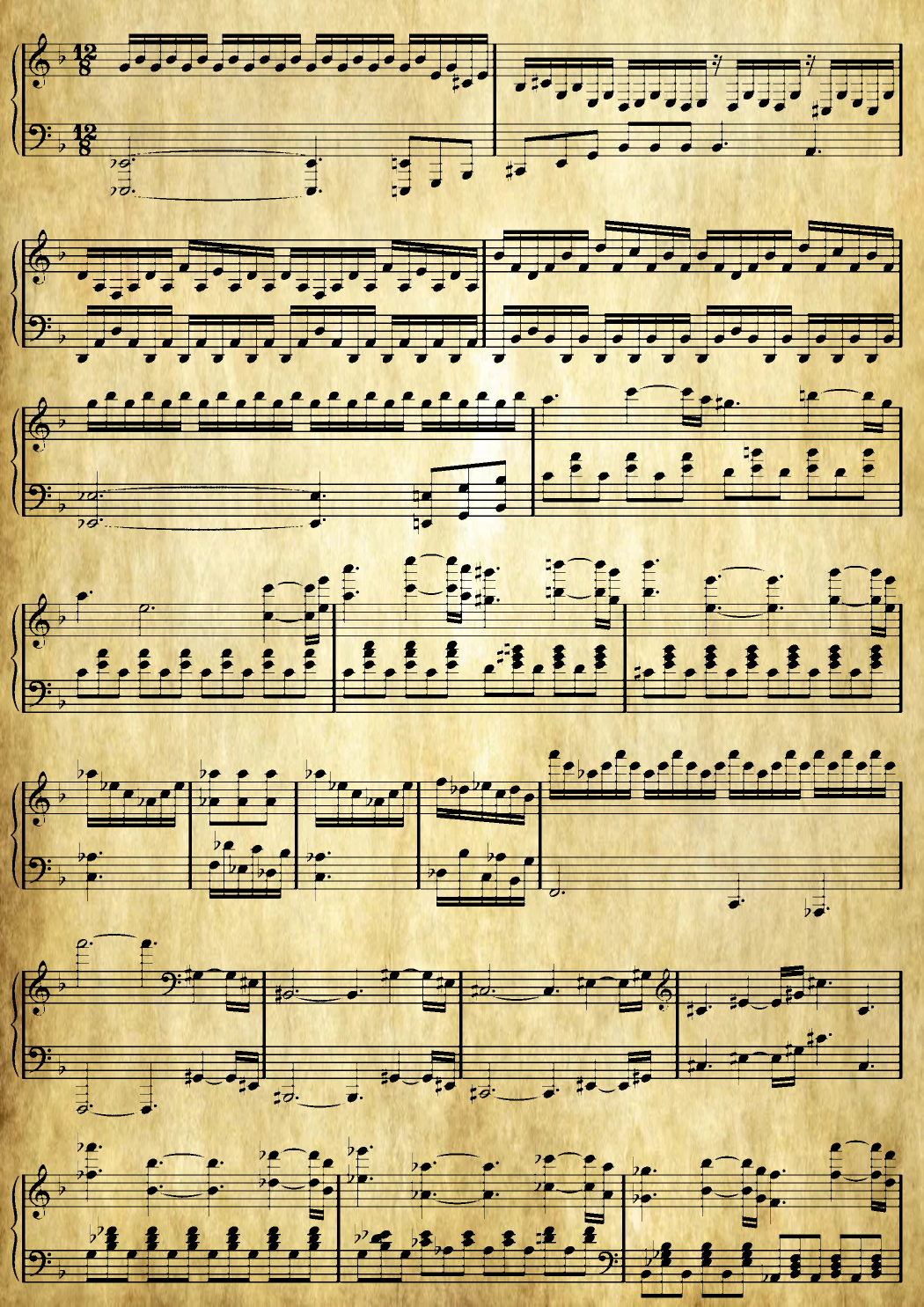}
         \caption{}
         \label{fig:sample_medium}
     \end{subfigure}
     \begin{subfigure}[b]{0.32\textwidth}
         \centering
         \includegraphics[width=\textwidth]{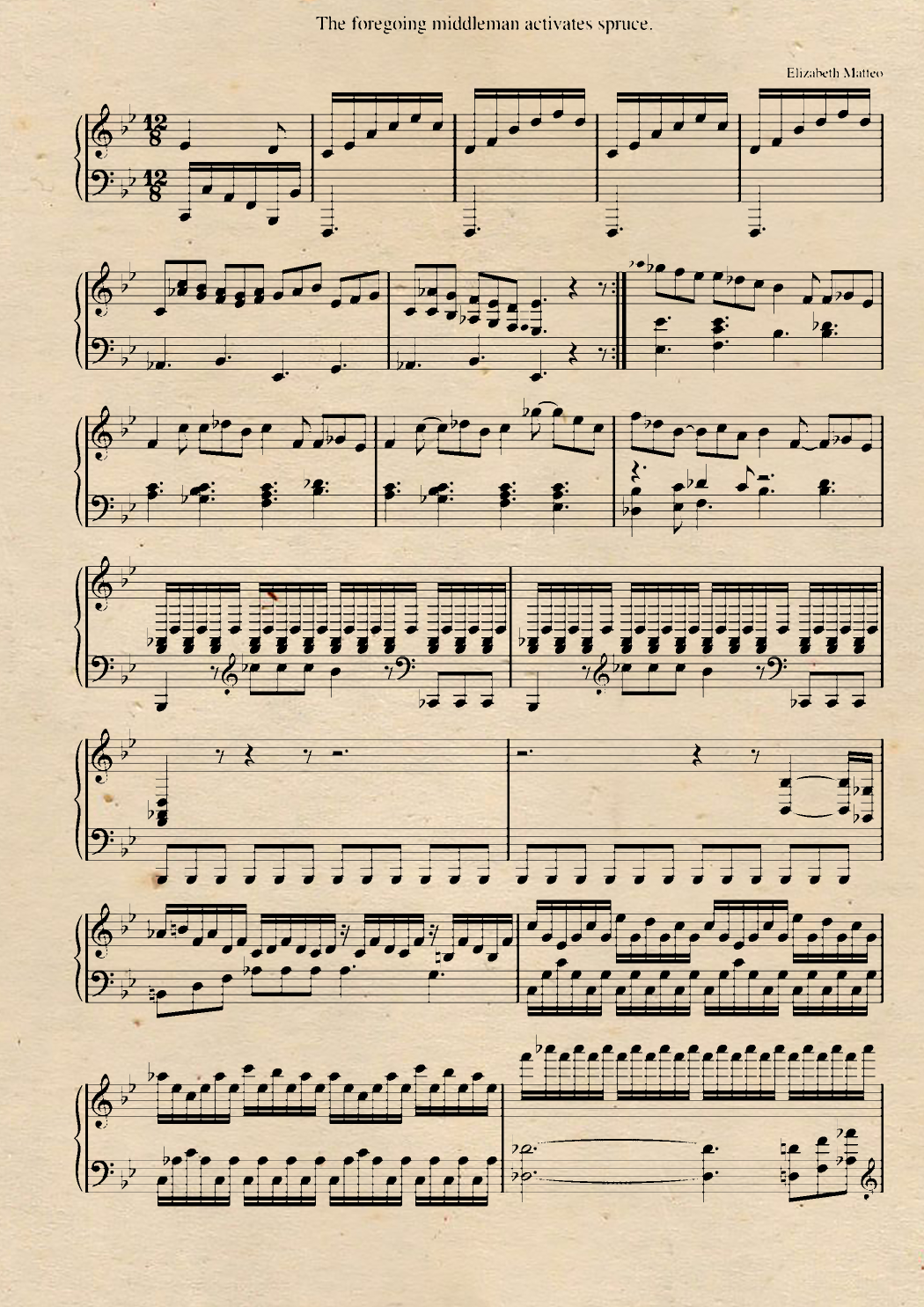}
         \caption{}
         \label{fig:sample_hard}
     \end{subfigure}
     \caption{\review{Three synthetic samples produced by the proposed generator. The music scores have been generated with different hyper-parameterization. Sample \ref{fig:sample_easy} is a basic four-systems page, sample \ref{fig:sample_medium} is a seven-system page with a random texture, and \ref{fig:sample_hard} is a seven-system page with a random texture, generated title and author, and random margin values.}}
     \label{fig:synthetic_samples}
\end{figure*}

Figure \ref{fig:synthetic_samples} shows three music scores generated by this algorithm. Note that this system can be run with any single-system score dataset annotated in \krn{}. In our case, it is applied to the generation of pianoform music scores.

\subsection{System-level Pre-training}
\label{sec:cl_first}
The first step of this training strategy is to pre-train the model to learn the first reading order complexity in full-page polyphonic scores: system-level transcription. Unlike, the \ac{OCR} and \ac{HTR} fields, this task is already complex, since polyphonic systems are composed of a group of staves that have to be simultaneously read in order to retrieve adequately their content. Given the \krn{} syntax, this means that the attention must interleave staves from bottom to top. This step is crucial to learn the transcription of full pages, since this first task is already complex~\citep{RiosVila:IJDARICDAR:2023, RiosVila:ICDAR:2024}.

\subsection{Incremental Curriculum Learning}
\label{sec:cl_second}
Once the \ac{SMT} is trained to transcribe single-system images, it undergoes a curriculum learning-based process that gradually transitions it from system-level reading to full-page transcription.

This stage is conducted using an incremental approach, where the number of systems on the pages increases linearly over the training steps. The generator starts by producing pages with two systems, since the network is already familiar with transcribing single-system samples, and gradually increases until a value of $n$ systems per page is reached. The number of systems per page increases after a specified number of training steps $s$. To avoid catastrophic forgetting, the generator does not consistently produce pages with a fixed number of systems, but instead randomly selects a number of systems from a range between two (the lowest difficulty) and the upper value $n$.

Both parameters, $n$ and $s$, are hyperparameters that must be determined empirically.

\subsection{Curriculum Learning-based Fine-tuning}
\label{sec:cl_third}
The final training step is to fine-tune the model on the target dataset. Therefore, a second curriculum learning stage is implemented. To balance the training process, the model is fed with both synthetic and real data samples using a probabilistic policy. The linear function described in Eq. \ref{eq:prob_func} determines the probability threshold $P$ for selecting either a synthetic or real sample:

\begin{equation}
\label{eq:prob_func}
\begin{split}
    P &= {\text{max}}(P_{{\text{min}}}, P_e) \\
    P_e &= P_{\text{max}} + \frac{s \cdot (P_{\text{min}} - P_{\text{max}})}{N}
\end{split}
\end{equation}

where $P$ is the probability threshold, $P_{\text{max}}$ and $P_{\text{min}}$ are the maximum and minimum probabilities of this threshold, $s$ is the current training step, and $N$ is the range over which the linear function is distributed. This scheduler transitions from providing mostly synthetic samples in the early stages to predominantly real samples when $s~\textgreater~N$.

\section{Experimental Setup}
\label{sec:experiments}
In this section, we describe the experimental environment designed to evaluate the performance of the proposed approach.\footnote{The implementation of the model is available at \url{https://github.com/antoniorv6/SMT} while the model and the weights are published in \url{https://huggingface.co/PRAIG}.}

\subsection{SMT Configuration}
Following the insights of previous research, we propose two different feature extraction approaches for the \ac{SMT}~\citep{RiosVila:ICDAR:2024}. The model consists of a Transformer decoder with eight layers, four attention heads, and an embedding size of $256$ features in both the attention and feed-forward modules. The first variation is the \FPSMTCNN{}, which implements a \review{Fully} \ac{CNN}. The backbone of this model rescales the height of the image by a factor of 16 and the width by a factor of 8 through pooling operations, as done in other recent works in the \ac{HTR} field~\citep{Singh:ICDAR:2021,Coquenet:TPAMI:2023,Dhiaf2023}. 
The second variation is the \FPSMTNXT{}, which is based on recent advancements in computer vision related to feature extraction networks. Specifically, we implement ConvNeXt~\citep{Liu:CVPR:2022}, an adaptation of the traditional ResNet50 that incorporates properties of the Swin Transformer~\citep{Liu:ICCV:2021}. It should be noted that we do not include the Swin Transformer itself, as previous experiments showed that it fails to converge, even in the simplest end-to-end \ac{OMR} scenarios~\citep{RiosVila:ICDAR:2024}. This is mainly because the data-devouring nature of the architecture, where the resources available are insufficient, as well as the lack of resolution of the available datasets. We train only the first three layers of the ConvNeXt backbone, with depths of 64, 128, and 256. \review{The network is trained from scratch with a learning rate of $10^{-4}$ and an early stopping criterion of five epochs in the three curriculum learning stages. To enhance the diversity and robustness of the training dataset, we implemented a data augmentation pipeline that includes random perspective transformations, elastic distortions, contrast adjustments, Gaussian blurring and random inversion. The model is also trained through teacher forcing, with a $20\%$ of noisy tokens, to avoid possible overfitting to the language model inputs.}

\subsection{Synthetic Generator Setup}
\begin{figure}[h]
     \centering
     \begin{subfigure}[b]{\columnwidth}
         \centering
         \includegraphics[width=\textwidth]{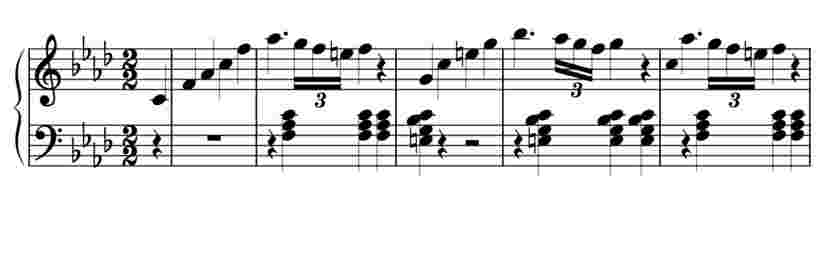}
         \caption{\review{Original sample.}}
         \label{fig:sample2}
     \end{subfigure}
     \begin{subfigure}[b]{\columnwidth}
         \centering
         \includegraphics[width=\textwidth]{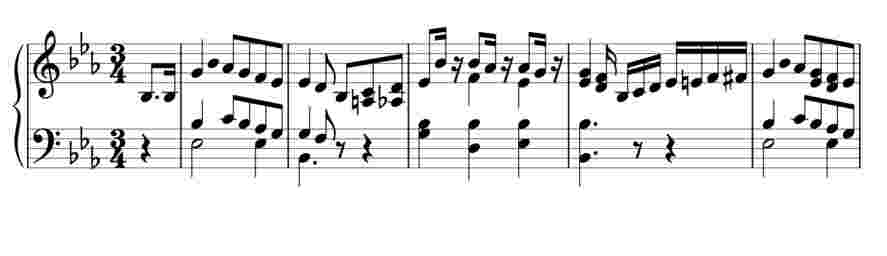}
         \caption{\review{Distorted sample found in the training set for synthetic generation.}}
         \label{fig:sample3}
     \end{subfigure}
    \caption{\review{Two GrandStaff samples from the Beethoven Sonatta piano n. 5.}}
     \label{fig:grandstaff_differences}
\end{figure}

In our study, the pool of music systems for generating images is sourced from the \GrandStaff{} dataset~\citep{RiosVila:IJDARICDAR:2023}, a public single-system music score dataset designed for end-to-end pianoform transcription. The generator uses \review{exclusively} the training set of the corpus, which consists of \review{deeply musically distorted scores to enhance the diversity and variability of both images and transcriptions}. \footnote{These distortions are applied through transposition, rythm and offset distortions.} This approach ensures that the generated data does not replicate any existing music scores, \review{as the distortions are applied to guarantee that no evident relationships are found between the original score and the distortion. Figure \ref{fig:grandstaff_differences}}. 

For the specific hyperparameters introduced in Section \ref{sec:cl_third}, the following values were used in all experiments: $n = 5$, $s = 40,000$, $P_{\text{max}}=0.9$, $P_{\text{min}}=0.2$, and $N=200,000$.

\subsection{Evaluation Protocol}
The assessment of OMR performance is one of the main topics of discussion in the field. Several works have extensively researched and proposed solutions to provide metrics that indicate the quality of these systems~\citep{Bellini:CMJ:2007, Byrd:JNMR:2015, Hajic:ISMIR:2016, Hajic:ICDAR:2017, mengarelli2020omr, Torras:Arxiv:2023}. Despite these efforts, there is no consensus on how to measure the \textit{goodness} of OMR systems, mainly due to specific issues surrounding the interpretation of music and the impact of errors on its comprehension~\citep{Calvo-Zaragoza:ACM:2020}.

In this work, we understand the OMR task as a pattern recognition problem, where the model's ability to recover the music notation depicted in the image is evaluated, rather than the quality of its eventual interpretation in musical terms. This non-semantic evaluation has been largely considered in recent literature~\citep{Calvo-Zaragoza:ISMIR:2018,Calvo-Zaragoza:PRL:2019,Baro:ICFHR:2022,Alfaro-Contreras:PRL:2022,RiosVila:ICDAR:2021,RiosVila:DLFM:2020,RiosVila:ISMIR:2022,AlfaroContreras:IJMIR:2023}. In this case, the performance of OMR is assessed using metrics based on the edit distance, which serves as a good proxy for the effort a user would need to invest in correcting the automatic transcription. This implicitly correlates with the quality of the transcription from a readability perspective.

In this work, we utilize three established metrics in the field for polyphonic transcription~\citep{RiosVila:IJDARICDAR:2023,Mayer:ICDAR:2024}, all of which are based on the normalized edit distance but differ in the basic units used for computing that distance. These metrics are the \acf{CER}, which evaluates among minimum semantic units in the output encoding; the \acf{SER}, a commonly used metric that takes complete symbols as evaluation tokens; and the \acf{LER}, which considers entire \krn{} lines as tokens, thus assessing the model's accuracy with regard to both content and document structure.\footnote{Evaluating the structure of the document is a key aspect in OMR when the transcription is more complex than the staff level, since the correct alignment between transcribed notes is crucial for retrieving accurate notation.} 

It should be noted that, regardless of the encoding used in the \ac{SMT}, all results are converted into \bkrn{} for computing the evaluation metrics. This ensures a fair comparison, where errors are equally penalized.

\section{Results}
\label{sec:results}
In this section, we evaluate the performance of the \ac{SMT} in two distinct scenarios: a controlled scenario using a rendered corpus of real music data (``FP-GrandStaff''), to study the best model configuration and tokenization method, and a real-case scenario using two scanned music corpora (``Mozarteum'' and ``Polish Digital Scores''), where we compare the \ac{SMT} with a commercial OMR software. This comparison allows us to assess both the challenges and the advantages of the proposed method. The dataset will be presented in their respective experimental scenario, although a general summary of their features is provided in Table \ref{tab:corpus-features}.

\begin{table*}[h]
\renewcommand{\arraystretch}{1.1}
\centering
\caption{Summary of the features of the data sources employed in this paper, such as image sizes in pixels, number of samples and staves per page, number of symbols present, and unique symbols (vocabulary).}
\label{tab:corpus-features}
\begin{tabular}{llllll}
\hline
 & \textbf{FP-GrandStaff} &  & \textbf{Mozarteum} &  & \textbf{Polish Digital Scores} \\ \hline
Num pages & 688 &  & 101 &  & 117 \\
Avg. page size & $2970 \times 2100$ &  & $1229 \times 900$ &  & $2100 \times 1484$ \\ \hline
Avg. systems per page & 5 &  & 6 &  & 6 \\
Max. systems per page & 7 &  & 7 &  & 7 \\
Min. systems per page & 1 &  & 5 &  & 3 \\ \hline
Avg. symbols per page &  &  &  &  &  \\
~ ~ \krn{} & 1184 &  & 1264 &  & 1303 \\
~ ~ \ekrn{} & 1363 &  & 1513 &  & 1621 \\
~ ~ \bkrn{} & 1897 &  & 2006 &  & 2196 \\
Max. symbols per page &  &  &  &  &  \\
~ ~ \krn{} & 2316 &  & 2156 &  & 2572 \\
~ ~ \ekrn{} & 3112 &  & 2461 &  & 3134 \\
~ ~ \bkrn{} & 4358 &  & 3264 &  & 4116 \\
Min. symbols per page &  &  &  &  &  \\
~ ~ \krn{} & 30 &  & 466 &  & 626 \\
~ ~ \ekrn{} & 30 &  & 430 &  & 865 \\
~ ~ \bkrn{} & 34 &  & 32 &  & 1103 \\ \hline
Unique symbols &  &  &  &  &  \\
~ ~ \krn{} & 20578 &  & 21515 &  & 24947 \\
~ ~ \ekrn{} & 5282 &  & 5327 &  & 5600 \\
~ ~ \bkrn{} & 184 &  & 191 &  & 215 \\ \hline
\end{tabular}
\end{table*}

\subsection{Controlled Scenario}
The different configurations of the \ac{SMT} are tested using the Full-Page GrandStaff (FP-\GrandStaff{}) dataset. This corpus consists of $688$ full-page music score images from six well-known composers, extracted from the Humdrum repository.\footnote{\url{https://github.com/humdrum-tools/humdrum-data}} The music score images were processed and rendered in the same manner as described by \cite{RiosVila:IJDARICDAR:2023}, as they represent the full-page versions of the scores provided in the \GrandStaff{} dataset. For our experiments, the FP-\GrandStaff{} corpus was divided into three partitions: 60\% of the samples for training, 20\% for validation, and 20\% for testing.

Table \ref{tab:results} reports the average \ac{CER}, \ac{SER}, and \ac{LER} obtained on the test set of the FP-\GrandStaff{} dataset for each network configuration and encoding method considered.

\begin{table*}[]
\renewcommand{\arraystretch}{1.1}
\centering
\caption{Average CER, SER, and LER (\%) obtained by the models studied in the test set of the FP-\GrandStaff{} dataset.}
\label{tab:results}
\begin{tabular}{@{}llcc@{}}
\toprule
\textbf{Encoding} & \textbf{} & \multicolumn{1}{l}{\textbf{\FPSMTCNN{}}} & \multicolumn{1}{l}{\textbf{\FPSMTNXT{}}} \\ \midrule
~~~~~\multirow{3}{*}{\krn{}} & CER & 15.8 & 7.2 \\
 & SER & 25.2 & 9.7 \\
 & LER & 44.3 & 19.9 \\ \hdashline
~~~~~\multirow{3}{*}{\ekrn{}} & CER & 17.3 & 6.8 \\
 & SER & 21.3 & 8.5 \\
 & LER & 35.7 & 16.6 \\ \hdashline
~~~~~\multirow{3}{*}{\bkrn{}} & CER & 15.2 & \textbf{5.6} \\
 & SER & 19.7 & \textbf{6.9} \\
 & LER & 32.0 & \textbf{12.9} \\ \bottomrule
\end{tabular}
\end{table*}

The results generally demonstrate that our approach for building and training the \ac{SMT} leads to successful end-to-end full-page transcription. Specifically, the \FPSMTNXT{} yields the best performance across all encodings considered, showing an average improvement of $62.1$\% in the \ac{SER} metric compared to its alternative. This confirms that modern convolutional approaches, despite being inspired by advancements in Vision Transformer architectures, are capable of learning appropriate features from complex music score documents.

From an encoding perspective, the two proposed encoding alternatives to the raw tokenization enhance the performance of the model. Both the \ekrn{} and the \bkrn{} encodings outperform raw \krn{}. Specifically, the \FPSMTNXT{} model reports that the \ekrn{} \ac{SER} improves results by $12.4\%$ and \bkrn{} by $28.6\%$. Among these alternative representations, the \bkrn{} notation yields a $18\%$ lower \ac{SER} than the \ekrn{} encoding, \review{following the insights got from previous works~\citep{RiosVila:IJDARICDAR:2023}}. This suggests that it is preferable to rely on a semantic-aware tokenization of the ground-truth, even if some correlation with the image is lost.

Based on these preliminary results, we claim that the \ac{SMT} is a viable end-to-end method for transcribing full-page pianoform documents. Among the configurations studied, the combination of \FPSMTNXT{} with \bkrn{} encoding proves to be the most effective for this task.

\subsubsection{Ablation Studies}
\label{sec:ablations}
In addition to the preliminary results provided above, we conducted several ablation studies in this controlled scenario to evaluate the relevance of key features of our approach. All experiments presented in this section were performed using the \FPSMTNXT{} model with the \bkrn{} notation, as this combination reported the best results. We assume that similar trends will hold for other configurations.

\paragraph{Ablation on Training Stages}
The first ablation study focuses on the training pipeline described in Section \ref{sec:cl}, specifically on the process of teaching the model the reading order of full pages. In this ablation study, we report the performance of the \FPSMTNXT{} model whether or not including two of the training stages: system-level pre-training and incremental curriculum learning. \review{The fine-tuning stage is considered mandatory for the model to converge on real music data}. The results of this ablation study are reported in Table \ref{tab:ablation1}.

\begin{table}[h]
\centering
\caption{Evaluation metrics of the model according to the ablation study on the training stages: (1) system-level pre-training, (2) incremental curriculum learning. \review{All of the models have been trained the same amount of epochs, evenly distributed between stages, to ensure an equal comparison.}}
\label{tab:ablation1}
\begin{tabular}{@{}lcclccc@{}}
\toprule
 & \multicolumn{2}{c}{\textbf{Steps}} &  & \multicolumn{3}{c}{\textbf{Metrics}} \\
 & \textbf{(1)} & \textbf{(2)} &  & \textbf{CER} & \textbf{SER} & \textbf{LER} \\ \midrule
\multirow{4}{*}{FP-GrandStaff} & \xmark & \xmark &  & 32.1 & 42.6 & 71.5 \\
 & \cmark & \xmark &  & 8.1 & 9.8 & 17.9 \\
 & \xmark & \cmark &  & 10.1 & 12.4 & 21.9 \\
 & \cmark & \cmark &  & \textbf{5.6} & \textbf{6.9} & \textbf{12.9} \\ \bottomrule
\end{tabular}
\end{table}

The individual analysis shows a significant improvement when performing at least one of the proposed stages, with a $77\%$ reduction in \ac{SER} when pre-training on single systems, and a $71\%$ reduction when using incremental curriculum learning. This indicates that both steps are crucial for achieving successful full-page transcription. More specifically, system-level pre-training is identified as the most impactful stage, allowing the model to understand the hierarchical reading order of pianoform documents step by step. After pre-training, the model only needs to learn how to sequentially read systems from top to bottom, which appears to be an easier task once single systems are mastered. Despite this, using all stages together results in superior performance compared to applying each step individually, with a relative improvement of up to $83\%$ over the baseline. This demonstrates that all stages of the training pipeline are essential for optimal performance.

\review{To further research on the effectiveness of the curriculum learning, we analyze the model performance across different page complexities during training. To do so, we generated a fixed test set from our synthetic generator ranging from one to eight systems per page.\footnote{\review{The excerpts used for testing were extracted from the test set of the \GrandStaff{} dataset, which were never seen during training.}} Table \ref{tab:curriculum_matrix} presents the performance matrix on this experiments.}

\begin{table*}[h]
\centering
\caption{\review{SER (in \%) performance matrix of the SMT during the curriculum learning stages. The rows indicate the maximum number of systems used during training, columns indicate number of systems in test pages.}}
\label{tab:curriculum_matrix}
\begin{tabular}{@{}c*{8}{c}@{}}
\toprule
\multirow{2}{*}{\review{\textbf{Max. Training Systems}}} & \multicolumn{8}{c}{\review{\textbf{No. Test Systems}}} \\
\cmidrule(l){2-9}
& \review{\textbf{1}} & \review{\textbf{2}} & \review{\textbf{3}} & \review{\textbf{4}} & \review{\textbf{5}} & \review{\textbf{6}} & \review{\textbf{7}} & \review{\textbf{8}} \\
\midrule
\review{\textbf{1}} & \review{5.1} & \review{48.7} & \review{67.4} & \review{78.9} & \review{85.3} & \review{89.6} & \review{92.4} & \review{94.8} \\
\review{\textbf{2}} & \review{5.2} & \review{12.2} & \review{45.6} & \review{62.1} & \review{73.8} & \review{81.5} & \review{87.2} & \review{91.6} \\
\review{\textbf{3}} & \review{4.4} & \review{8.8} & \review{13.2} & \review{47.2} & \review{59.7} & \review{68.9} & \review{76.4} & \review{82.7} \\
\review{\textbf{4}} & \review{4.1} & \review{5.0} & \review{7.1} & \review{12.3} & \review{19.8} & \review{25.3} & \review{52.7} & \review{63.9} \\
\review{\textbf{5}} & \review{4.2} & \review{4.91} & \review{6.0} & \review{8.24} & \review{13.9} & \review{18.9} & \review{32.5} & \review{35.3} \\
\bottomrule
\end{tabular}
\end{table*}

\review{
The matrix shows interesing insights about the learing process of the model. The \FPSMTNXT{} is intended to learn two things: (i) inferring the correct reading order across multiple systems, and (ii) maintaining accurate symbol classification. The curriculum learning is designed to specifically learn the first skill, while maintaining some accuracy in the second one. The dramatic performance degradation when test complexity exceeds training complexity, e.g., SER jumping from 5.1\% to 48.7\% when a single-system model encounters 2-system pages, demonstrates that--in fact-- this skill is polished during this process. Conversely, models trained on higher complexity show robust generalization---a model trained on 5 systems maintains SER below 9\% for all pages with 1--4 systems.
The evaluation on out-of-distribution test cases---with pages containing 6--8 systems, exceeding the maximum training complexity---provides critical evidence for the generalization capabilities of the model. The observed performance degradation follows a sub-linear pattern, with SER increasing from 15.9\%, on 5 systems, to 35.3\%, rather than exhibiting catastrophic failure. This controlled degradation suggests that the model has internalized generalizable representations for spatial layout understanding and sequential reading order, rather than overfitting to specific page configurations. These findings validate our hypothesis that incremental curriculum learning establishes a robust initialization for the subsequent fine-tuning stage, where the pre-trained reading order comprehension allows the optimization process to concentrate on domain adaptation and symbol classification refinement.}

\paragraph{Ablation on Positional Encoding}
Another key aspect of the proposal is the \ac{PE} applied to transfer the information from the feature map, provided by the backbone, to the Transformer decoder. For this purpose, we use the 2D \ac{PE} formula, as defined in Equation \ref{eq:PE2D} in Section \ref{sec:decoder}. However, Transformers were originally designed to use a 1D \ac{PE} setting~\citep{Kim:ECCV:2022,Blecher:Arxiv:2023,Li:Arxiv:2021}. This ablation study analyzes the impact of employing the 2D \ac{PE} formula in this task and assesses the performance difference between using a 2D setting versus a 1D one.

\begin{table}[h]
\centering
\renewcommand{\arraystretch}{1.2}
\caption{Evaluation metrics of the model according to the ablation study on Positional Encoding.}
\label{tab:ablation2}
\begin{tabular}{@{}llccc@{}}
\toprule
\multirow{2}{*}{\textbf{Position Encoding}} & \textbf{} & \multicolumn{3}{c}{\textbf{FP-\GrandStaff{}}} \\
 & \textbf{} & \textbf{CER} & \textbf{SER} & \textbf{LER} \\ \midrule
No Position Encoding &  & 57.3 & 78.4 & 100 \\
Position Encoding 1D &  & 13.6 & 15.57 & 23.7 \\
Position Encoding 2D &  & 5.6 & 6.9 & 12.9 \\ \bottomrule
\end{tabular}
\end{table}

Table \ref{tab:ablation2} shows the results obtained in three scenarios: no \ac{PE}, 1D \ac{PE}, and 2D \ac{PE}. The baseline results, where no positional information is provided, indicate that applying \ac{PE} to the feature map is essential for successful full-page transcription. This outcome was rather expected. Specifically, the 2D \ac{PE} yields superior results, outperforming the 1D formula by $55.7\%$ in the \ac{SER} and $45.5\%$ in the \ac{LER}. These results demonstrate that providing 2D positional information not only improves transcription quality but also generates significantly more consistent documents. This is because the reading order in pianoform scores heavily depends on both axes. The explicit encoding in the 2D \ac{PE} helps the model identify the elements along the height of the document. In the case of 1D \ac{PE}, this information must be learned implicitly. Therefore, for transcribing complex full-page music scores, it is preferable to apply a 2D \ac{PE} formula, which aligns with insights reported in related fields~\citep{Singh:ICDAR:2021,Coquenet:TPAMI:2023}.

\subsection{Evaluation on Real Corpora}
For the experiments with real corpora, we tested the \FPSMTNXT{} model on two scanned score datasets. The first is the Digital Mozarteum Edition repertory, referred to as the Mozarteum dataset~\citep{Rink:NCMR:2011}. This dataset contains $101$ pages from a re-edition of the works of Wolfgang Amadeus Mozart, which are publicly available through the MoVi tool.\footnote{\url{https://dme.mozarteum.at/movi/en}} The second corpus is the Polish Digital Scores repository, which contains transcriptions from the \textit{Heritage of Polish Music in Open Access} project by the Chopin Institute.\footnote{\url{https://github.com/pl-wnifc/humdrum-polish-scores}} We selected 117 pianoform pages from this repository.

\begin{figure*}
     \centering
     \begin{subfigure}[b]{0.4\textwidth}
         \centering
         \includegraphics[width=\textwidth]{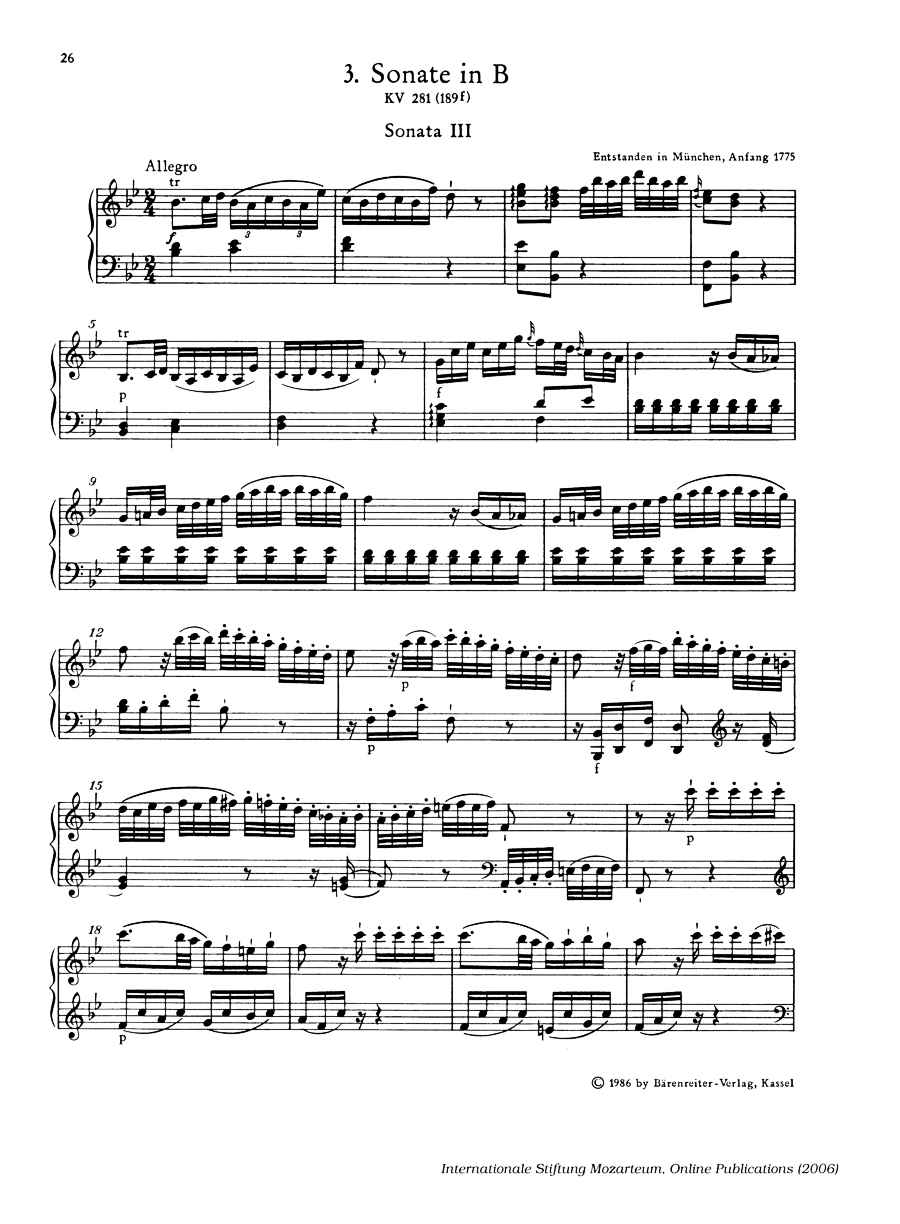}
         \caption{Mozarteum}
         \label{fig:sample2}
     \end{subfigure}
     \begin{subfigure}[b]{0.4\textwidth}
         \centering
         \includegraphics[width=\textwidth]{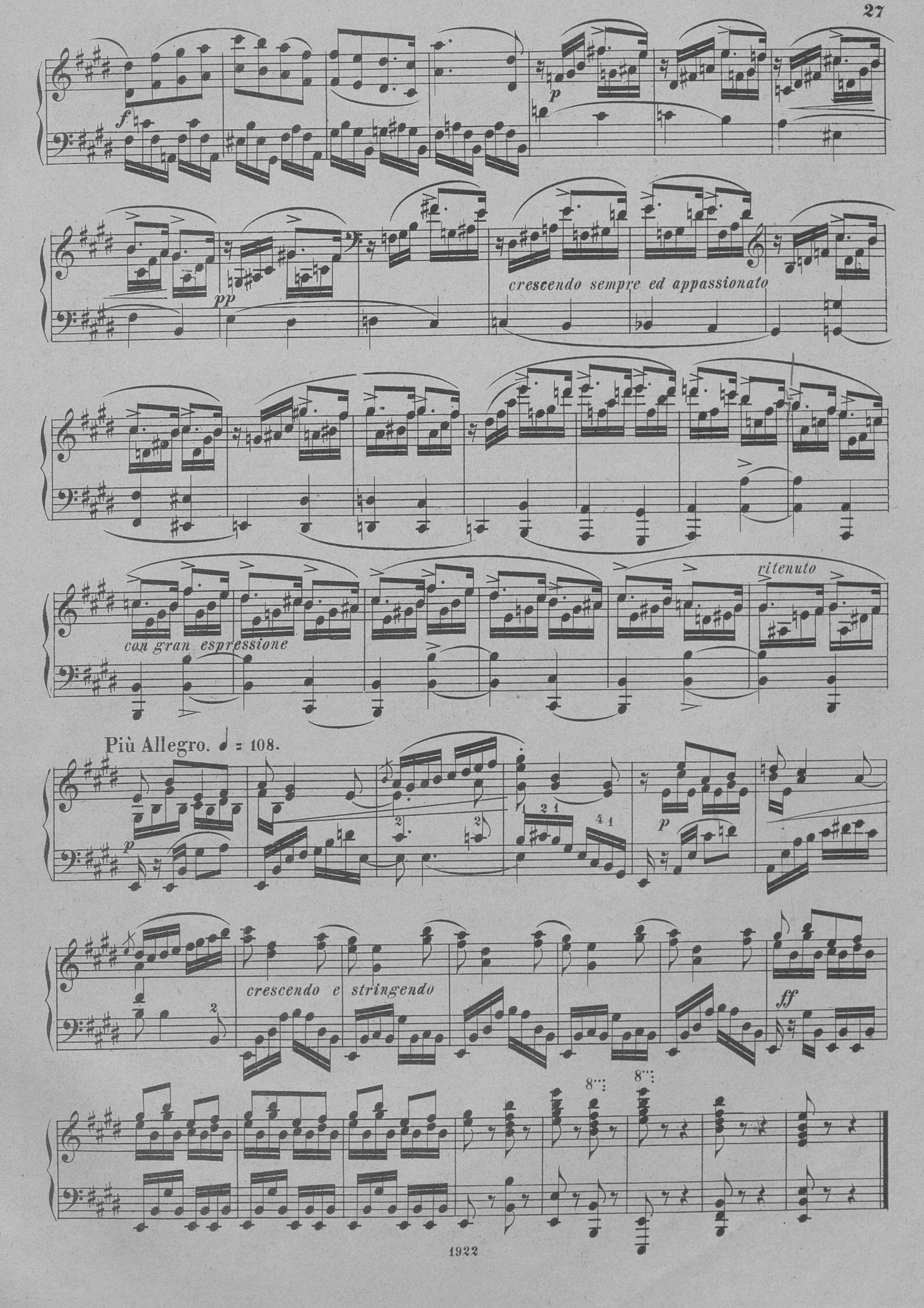}
         \caption{Polish Digital Scores}
         \label{fig:sample3}
     \end{subfigure}
     \caption{Examples from the corpora used in the real-case scenario evaluation.}
     \label{fig:real_sets}
\end{figure*}

Both corpora are initially stored as book-like PDFs, where multiple pages are found in a single file. We preprocessed them by separating the pages along with their corresponding transcriptions. An example from each dataset is shown in Fig. \ref{fig:real_sets}.

Since both datasets contain a limited number of samples, we used a 5-fold cross-validation\review{, where folds were created at the piece level to ensure no samples from the same musical work appear in both training and test sets}. Results are consistently reported by averaging the performance on the test set of each fold.

\review{\subsubsection{Baseline approaches}}
\label{subsec:commercial_presentation}
Since this is the first time an end-to-end deep learning model has been proposed for full-page \ac{OMR} in pianoform transcription, there are no similar publicly available solutions to compare our results with. To address this and provide a \review{comprehensive} evaluation, \review{we test different available \ac{OMR} tools and adaptations of state of the art approaches to address this task}.

\review{\paragraph{Commercial software}}
\review{First, we tested the datasets using commercial OMR software, which is, in fact, the tool a user would currently employ to transcribe these works.} Specifically, we transcribed the sources using the \review{PhotoScore tool, one of the most popular OMR softwares, which has also been used as a benchmark in related works~\citep{CalvoZaragoza:APPSCI:2018,Baro:GRCTE:2018,Baro:PRL:2019,DeGroot:ISMIR:2020} and SoundSlice, which is, to the best of our knowledge, one of the most recent deep learning-based OMR tools that is gaining increasing popularity by the community.\footnote{\url{https://www.soundslice.com/}}} Although \review{their} operation is not publicly documented, \review{these tools are believed to transcribe music at the full-page level}. \review{More details about the transcription pipeline with these tools is given in Appendix \ref{app:pipeline}.}

\review{\paragraph{Open-source software}}
\review{In this paper, we also feature two well-known open-source projects projects, which stand as free alternatives to commercial software. The first one is Audiveris, which is the open source counterpart of the PhotoScore tool that has a significant popularity over the OMR community.\footnote{\url{https://audiveris.en.lo4d.com/windows}} The second one is OEMER, which is a recent \ac{OMR} system that utilizes deep learning models to transcribe images of printed sheet music~\citep{Yoyo:Software:2023}. The transcription pipeline of these tools is the same described in Appendix \ref{app:pipeline}, since they export their results in the same format.}

Because the ability of the \ac{SMT} to be fine-tuned is a significant advantage, its comparison with a zero-shot application may be unfair. Therefore, we present two sets of results for the \FPSMTNXT{}: one where the model is trained only on synthetic data and another where the model is fine-tuned with the training partitions of the real scores, hereinafter referred to as \FPSMTNXTff{}.

\begin{figure*}
    \centering
    \includegraphics[width=\linewidth]{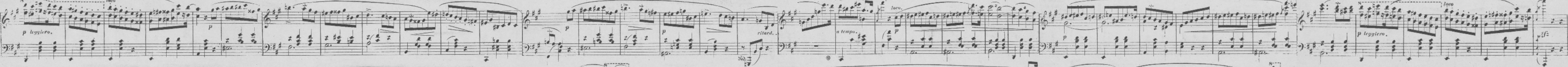}
    \caption{\review{Example of a composed score from the Polish Scores dataset to train the system-level transcriptors. Music systems were extracted with the YOLOv8 network from~\cite{Dvorak:WORMS:2024}}}
    \label{fig:composition}
\end{figure*}

\review{\paragraph{Layout Analysis \& Composition OMR}}
\review{The presented tools in this setup, although useful when dealing with the task with no resources, roughly represent the state of the art of \ac{OMR}. Typically, if provided with resources, such a project would be addressed through a two-stage pipeline compoded of a Layout Analysis that detects the regions and a system-wise transcription model. However, the Mozarteum and Polish Digital Scores datasets only provide the music score images and their correspondent full-page transcription. To provide a fair baseline, we resort to region composition~\citep{Tanha:ICDAR:2015,Villarreal:ICDAR:2024}. This approach consists of concatenating all the regions extracted from a generic Layout Analysis model and training a line-level transcription model with these images and the full-page ground-truth. This would be the most viable option to perform end-to-end \ac{OMR} without requiring manual relabeling of the datasets. We leverage the YOLO-v8 implemented by ~\cite{Dvorak:WORMS:2024} as the Layout Analysis backbone, which works very precisely on the Mozarteum and Polish Scores datasets, and set up three different system-level transcription models. The first one is the CTC-based unfolding network from \cite{RiosVila:IJDARICDAR:2023}, which represents a cheap alternative approach for system-wise transcription. Then, we leverage the \FPSMTNXTff{} and the TrOMR network, which is another transformer-based image-to-sequence architecture that is able to deal with polyphonic music~\citep{Li:ICASSP:2023}. An example of these composed scores is shown in Fig. \ref{fig:composition}. To assess fair comparisons between these fine-tuned models and the full-page \FPSMTNXTff{}, all of the models are pre-trained to transcribe single systems on the \GrandStaff{} dataset.}

\subsubsection{Results Analysis}
In Table \ref{tab:results_real}, we report the average \ac{SER} across the test folds, as well as the range of metric values for individual samples.

% \usepackage{booktabs}
\begin{table*}[h]
\centering
\renewcommand{\arraystretch}{1.4}
\caption{\review{Average evaluation metrics (in \%) and standard deviations for the evaluated datasets and methods in the real scores scenario. Bold values indicate the best overall performance, while underlined values highlight the best zero-shot performance. ``Success \%'' refers to the proportion of samples that did not cause the corresponding tool to crash.}}
\label{tab:results_real}
\resizebox{\textwidth}{!}{\begin{tabular}{@{}llcccccccc@{}}
\toprule
\multicolumn{1}{l}{\review{\textbf{Model}}} &  & \multicolumn{4}{c}{\review{\textbf{Mozarteum}}} & \multicolumn{4}{c}{\review{\textbf{Polish Scores}}} \\ \cmidrule(lr){3-6} \cmidrule(lr){7-10}
 &  & \review{\textbf{CER}} & \review{\textbf{SER}} & \review{\textbf{LER}} & \review{\textbf{Success \%}} & \review{\textbf{CER}} & \review{\textbf{SER}} & \review{\textbf{LER}} & \review{\textbf{Success \%}} \\ \midrule
\textit{\review{Commercial Software}} &  &  &  &  &  &  &  &  &  \\
~~~\review{PhotoScore} &  & \review{$66.3 \pm 10.2$} & \review{$70.3 \pm 12.4$} & \review{$100.0 \pm 15.0$} & \review{50\%} & \review{$73.3 \pm 16.3$} & \review{$83.3 \pm 16.3$} & \review{$100.0 \pm 18.1$} & \review{75\%} \\

~~~\review{SoundSlice} &  & \review{$31.3 \pm 2.4$}  & \review{$\underline{34.5 \pm 2.7}$} & \review{$\underline{50.1 \pm 3.7}$}  & \review{$100\%$} & \review{$\underline{43.1 \pm 8.5}$}  & \review{$\underline{54.9 \pm 11.1}$}  & \review{$\underline{66.6 \pm 9.6}$}  & \review{$100\%$} \\
 \hdashline
\textit{\review{Open source tools}} &  &  &  &  &  &  &  &  &  \\
~~~\review{Audiveris} &  & \review{$86.8 \pm 5.5$}  & \review{$94.6 \pm 4.0$}  & \review{$100.0 \pm 0.0$}  & \review{$96\%$} & \review{$93.7 \pm 0.0$}  & \review{$94.9 \pm 8.2$}  & \review{$100.0 \pm 8.0$}  & \review{$86\%$} \\
~~~\review{Oemer} & & \review{$85.1 \pm 6.2$} & \review{$90.3 \pm 6.8$} & \review{$100.0 \pm 0.0$} & \review{100\%} & \review{$68.0 \pm 1.3$} & \review{$75.2 \pm 2.2$} & \review{$100.0 \pm 0.0$} & \review{95\%} \\
\hdashline
\textit{\review{Layout Analysis \& Composition}} &  &  &  &  &  &  &  &  &  \\
~~~\review{Unfolding CRNN} &  & \review{$46.5 \pm 7.9$} & \review{$55.9 \pm 9.5$} & \review{$91.0 \pm 5.0$} & \review{100\%} & \review{$68.0 \pm 1.3$} & \review{$75.2 \pm 2.2$} & \review{$100.0 \pm 5.0$} & \review{100\%} \\

~~~\review{System-level \FPSMTNXTff{}} &  & \review{$20.6 \pm 3.1$} & \review{$21.5 \pm 4.0$} & \review{$39.3 \pm 2.9$} & \review{100\%} & \review{$32.1 \pm 3.3$} & \review{$33.7 \pm 4.9$} & \review{$64.8 \pm 3.6$} & \review{100\%} \\
~~~\review{TrOMR} &  & \review{$25.9 \pm 2.5$} & \review{$25.7 \pm 2.7$} & \review{$44.9 \pm 3.6$} & \review{100\%} & \review{$48.72 \pm 3.3$} & \review{$53.2 \pm 2.7$} & \review{$88.0 \pm 3.8$} & \review{100\%} \\ \hdashline
\textit{\review{Proposal}} &  &  &  &  &  &  &  &  &  \\
~~~\review{\FPSMTNXT{}} &  & \review{$\underline{27.7 \pm 2.0}$} & \review{$38.9 \pm 2.1$} & \review{$70.8 \pm 3.0$} & \review{100\%} & \review{$53.2 \pm 7.5$}  & \review{$65.4 \pm 11.3$} & \review{$78.9 \pm 10.5$}  & \review{$100\%$}  \\
~~~\review{\FPSMTNXTff{}} &  & \review{$\mathbf{10.5 \pm 3.8}$} & \review{$\mathbf{14.1 \pm 3.9}$} & \review{$\mathbf{28.2 \pm 3.6}$} & \review{100\%} & \review{$\mathbf{19.6 \pm 2.9}$} & \review{$\mathbf{25.8 \pm 5.2}$} & \review{$\mathbf{58.2 \pm 3.2}$} & \review{$100\%$} \\
\bottomrule
\end{tabular}}
\end{table*}

%\begin{table*}[h]
%\renewcommand{\arraystretch}{1.2}
%\centering
%\caption{Average SER (\%) and error range for the evaluated datasets in the real scores scenario.}
%\label{tab:results_real}
%\begin{tabular}{llccc}
%\hline
%\textbf{Model} & \textbf{} & \multicolumn{1}{l}{\textbf{Mozarteum}} & \multicolumn{1}{l}{\textbf{}} & \multicolumn{1}{l}{\textbf{Polish Digital Scores}} \\ \hline
%\multirow{2}{*}{\FPSMTNXT{}} & Avg. SER & 33.1 &  & 58.1 \\
% & Range & {[15, 67.5]} &  & {[31.2, 100]} \\
%\multirow{2}{*}{\FPSMTNXTff{}} & Avg. SER & 8.3 &  & 16.6 \\
% & Range & {[1.7, 40.3]} &  & {[7.9, 44.6]} \\ \hdashline
% \multirow{2}{*}{PhotoScore} & Avg. SER & 55.3 &  & 68.5 \\
% & Range & {[39.7, 84.2]} &  & {[30.5, 100]} \\ \hline
%\end{tabular}
%\end{table*}

\begin{figure*}
     \centering
     \begin{subfigure}[b]{\columnwidth}
         \centering
         \includegraphics[width=\textwidth]{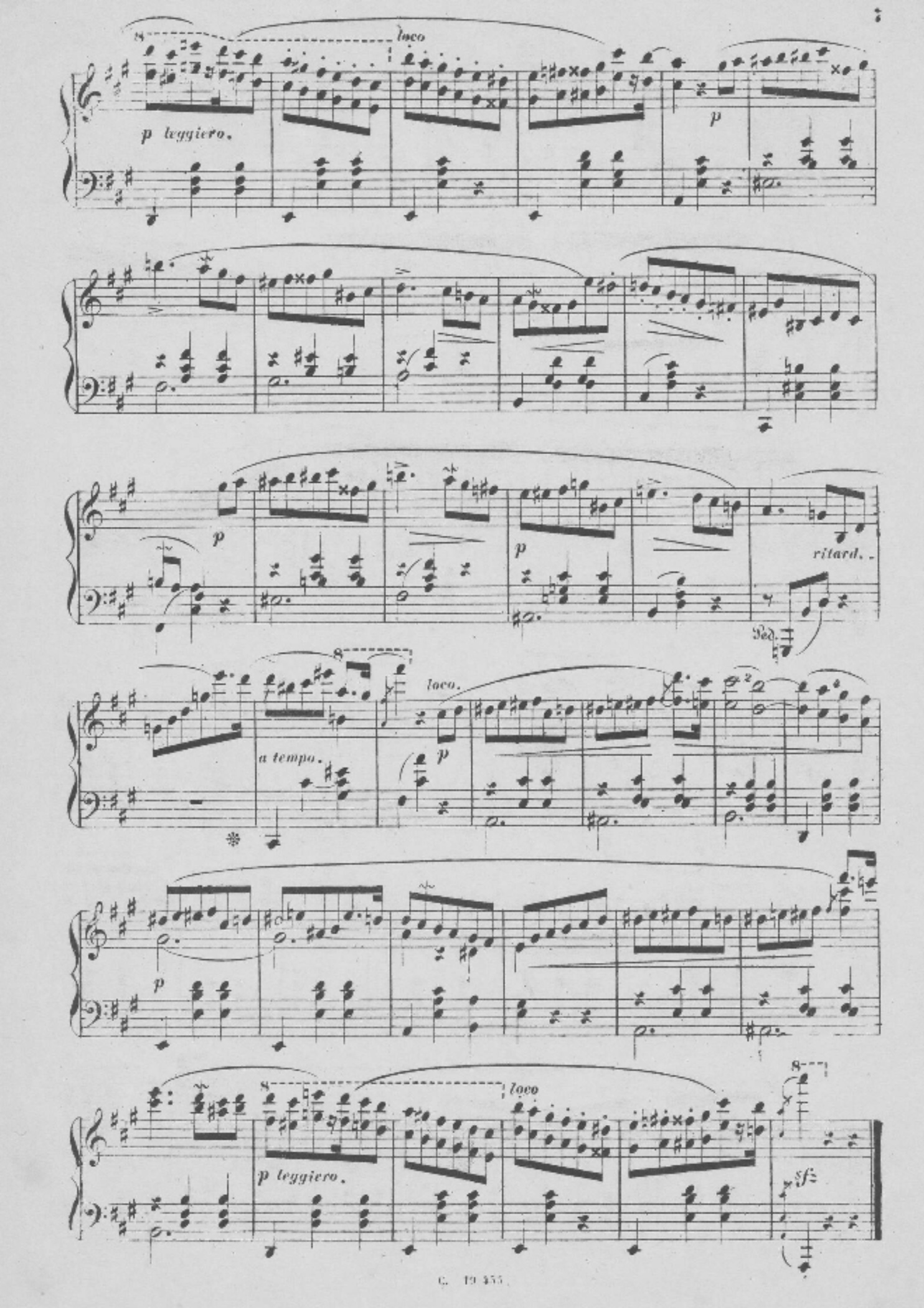}
         \caption{Input test sample}
         \label{fig:sample2}
     \end{subfigure}
     \begin{subfigure}[b]{\columnwidth}
         \centering
         \includegraphics[width=\textwidth]{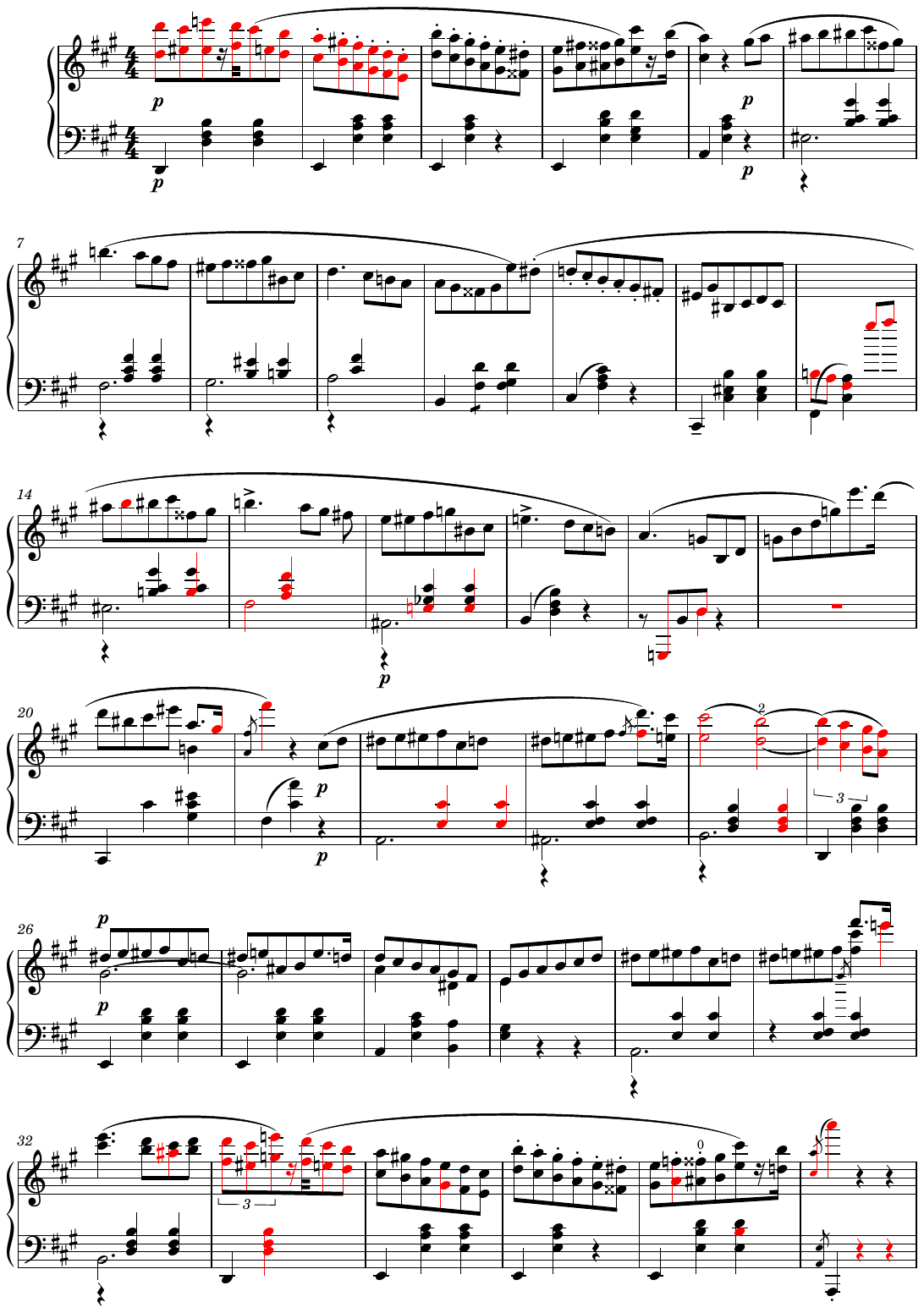}
         \caption{\review{SoundSlice}}
         \label{fig:sample3}
     \end{subfigure}
     \begin{subfigure}[b]{\columnwidth}
         \centering
         \includegraphics[width=\textwidth]{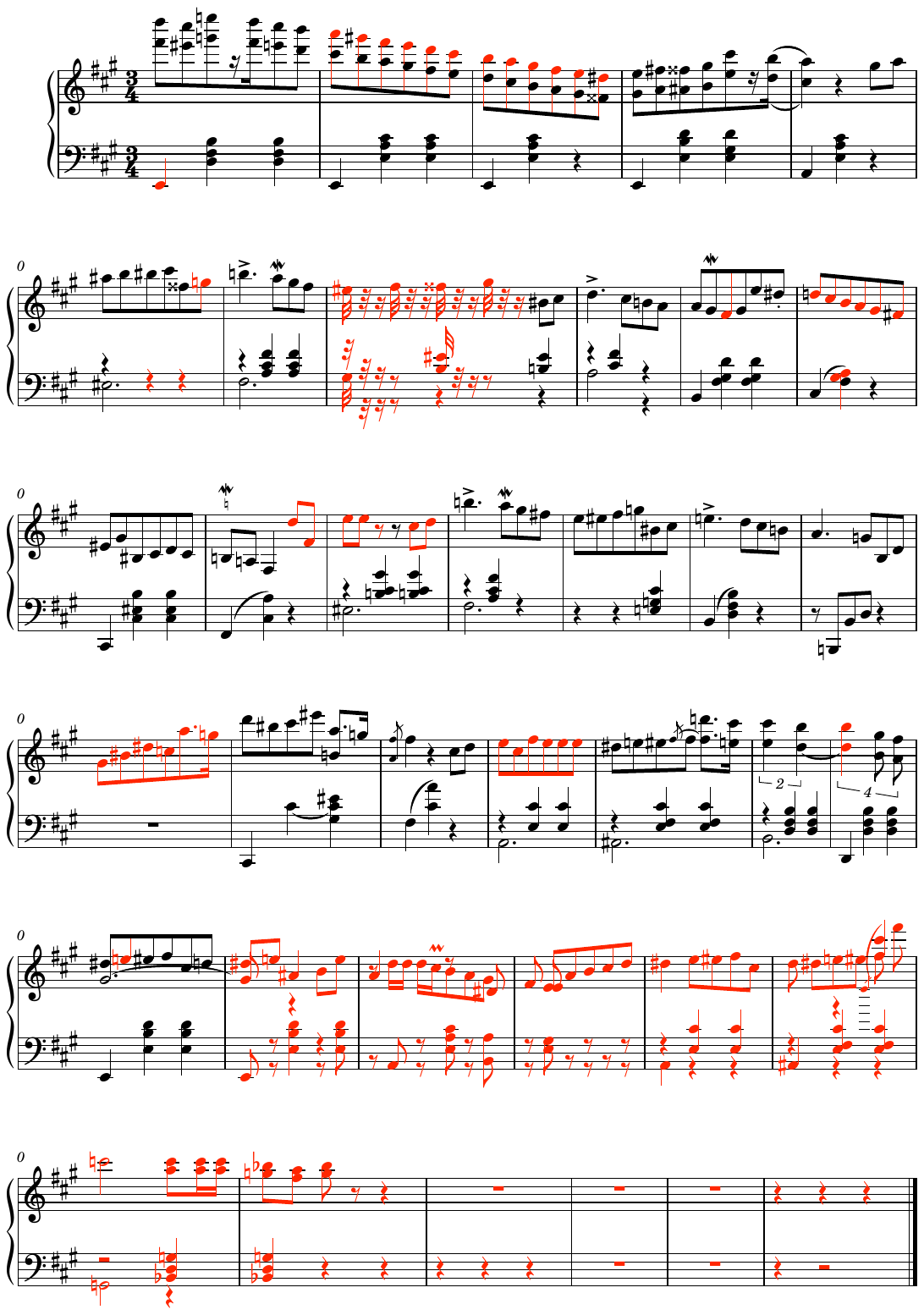}
         \caption{\FPSMTNXT{}}
         \label{fig:sample3}
     \end{subfigure}
     \begin{subfigure}[b]{\columnwidth}
         \centering
         \includegraphics[width=\textwidth]{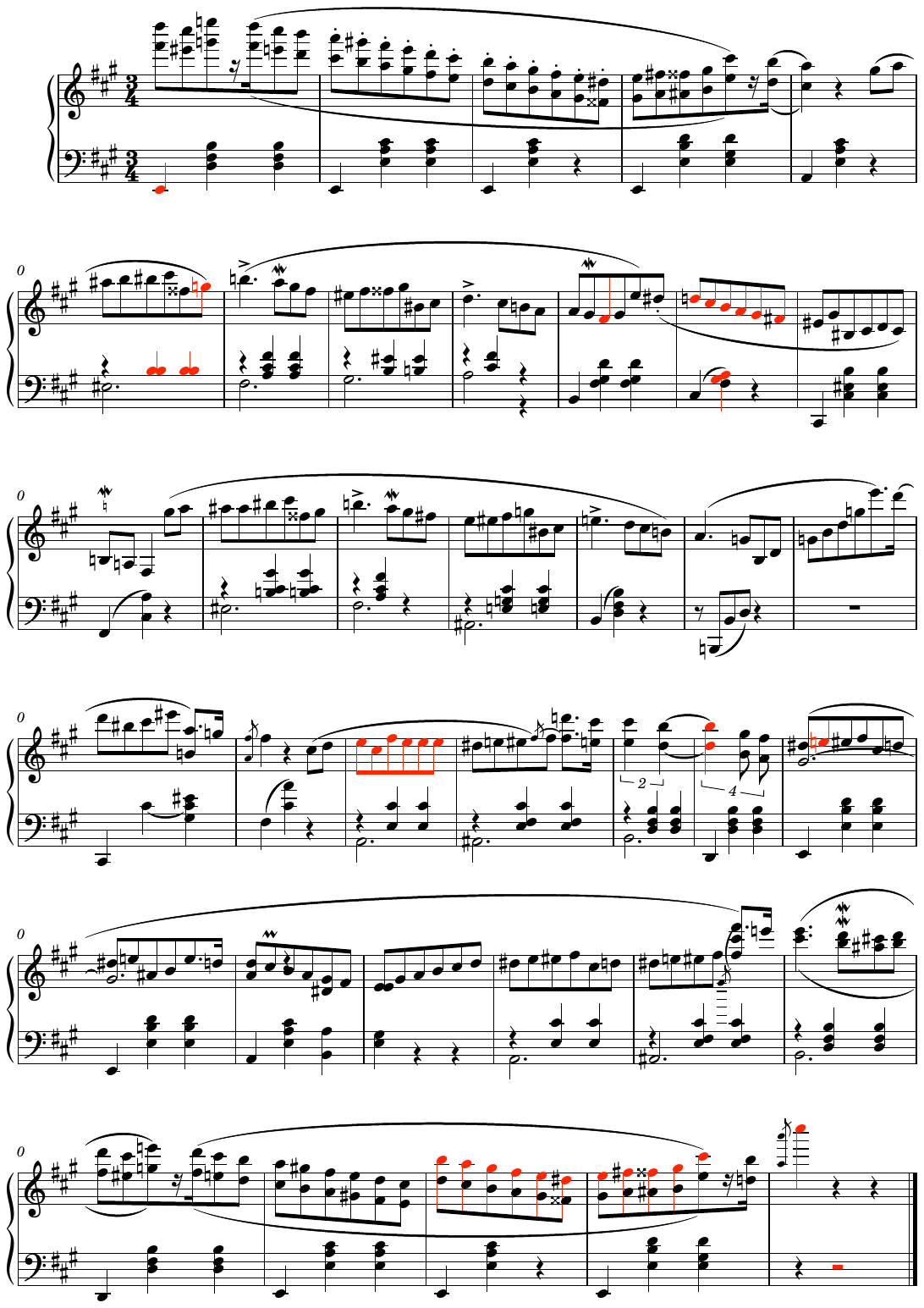}
         \caption{\FPSMTNXTff{}}
         \label{fig:sample3}
     \end{subfigure}
    \caption{\review{Visualization of the transcription performed by (b) SoundSlice, (c) the \FPSMTNXT{} and (d) the \FPSMTNXTff{} on a test sample (a) from the Polish Digital Scores corpus with highlighted errors. The visual evaluation was produced at the notation level. Note that some differences between the score and the renderings are not highlighted as errors, as the rendering tool may produce visually different outputs that are interpreted the same in the source score, as well as some errors in the output encoding are errors. This is specially seen in the case of SoundSlice, where note deletions, which is the greatest source of errors, are not represented accurately. The SER of each transcription is (b) 30.5\%, (c) 55.3\% and (d) 12.2\%.}}
     \label{fig:visual_results}
\end{figure*}

\review{First, we would like to emphasize that the results provided by the \ac{OMR} tools---PhotoScore, SoundSlice, Audiveris, and OEMER---demonstrate that the corpora used in this work are challenging. This is especially evident in the Polish Digital Scores scenario, where all tools dramatically fail in transcription accuracy. SoundSlice performs the best, being able to transcribe all of the test set scores with fair accuracy. Under the same conditions, the \FPSMTNXT{} reports similar results as SoundSlice. In the Mozarteum dataset, which has cleaner typeset images, the \FPSMTNXT{} yields an overall $5\%$ performance drop in comparison to the tool, except for the \ac{LER} metric, where the difference is more significant. In the Polish Scores Dataset, reported results are similar. Nevertheless, these results seem not to be significant, given the volatility of \ac{SER} shown by the deviation metrics. However, this comparison is meant to prove the difficulty of the challenge, as well as show the consistency Curriculum Learning gives to the \FPSMTNXT{} which reaches similar results as the best performant commercial OMR option.}

\review{When being able to fine-tune to the target domain, all methods show stronger performance, with the proposed \FPSMTNXTff{} consistently outperforming other state-of-the-art approaches across all evaluation metrics. Concerning the Layout Analysis \& Composition methods, we observe that the CTC-based unfolding approach---which yields even worse results than the best-performing commercial OMR tool---does not hold to the challenge of transcribing full-page scores, even adapting the problem with system-like structures. Regarding the image-to-sequence approaches, the \ac{SMT} outperforms the TrOMR approach, improving its performance by $16\%$ and $36\%$ in Mozarteum and Digital Polish Scores, respectively. This means that the \ac{SMT}, under the same conditions, is better at transcribing music notation than other Transformer-based approaches.}

\review{The proposed segmentation-free \FPSMTNXTff{} presents the best overall results in comparison to the Layout Analysis \& Composition systems, presenting a $14.1\%$ \ac{SER} in Mozarteum and a $25.9\%$ in Digital Polish Scores. These results, in comparison to the best state of the art, show an improvement of approximately $33\%$ and $23\%$. This discrepancy in fine-grained accuracy metrics demonstrates the advantages of holistic, end-to-end layout modeling. The superior performance of \FPSMTNXTff{} can be attributed to its ability to jointly model layout and content without decomposing the page into smaller units, thus better capturing long-range dependencies and structural regularities. In contrast, the system-level decomposition used in the alternative variant may introduce segmentation boundaries that hinder the capture of full-page context.}

\review{Notably, these results are achieved with fewer than 100 fine-tuning samples. This demonstrates that the pre-training strategies provide a solid foundation for understanding how music scores are read, while the fine-tuning process allows the system to learn details that synthetic and rendered data do not capture. These missing details may arise from two different sources: (i) the input image, which can be very specific to each corpus, and (ii) the number of missing tokens from the target dataset in the synthetic data.}

\review{To complement these results with a qualitative analysis, Fig. \ref{fig:visual_results} provides a visual comparison of the outputs from SoundSlice, the best-performant OMR tool, the \FPSMTNXT{} and the \FPSMTNXTff{} with the highlighted errors, marked with the \textit{musicdiff} tool~\citep{Foscarin:DLFM:2019}. First of all, we observe that in the \FPSMTNXT{} (b), all the voices are correctly placed, since its training is specific for pianoform scores. However, note that all slurs are skipped, which is visually impactful. This is because the slur token is one of the missing symbols of the \GrandStaff{} dataset, proving the performance downgrade. We also observe that most of the errors are concentrated in the last systems of the score, as the Curriculum Learning may not have seen enough samples of that complexity. Some symbols avoidance and misplacements are also observed, especially the dots---which are completely skipped. The \FPSMTNXTff{} result (d) shows that fine-tuning makes a very significant difference with other results, being able to catch the vocabulary divergences, as well as increasing the precision of the system to retrieve the content of the score, where only note misplacement is observed.}

\section{Conclusions}
\label{sec:conclusions}
\review{In this paper, we presented the first true end-to-end methodology for Optical Music Recognition (OMR) in complex pages}, which has traditionally relied on pipeline-based systems to perform full-page transcription. Our model is based on feature extraction and an autoregressive Transformer decoder for language modeling, where a full-page score is received as input and its transcription is provided directly in a music encoding format. This model is trained using comprehensive strategies, including system-level pre-training and curriculum learning.

We evaluated our approach in a full-page polyphonic music scenario, focusing on transcribing pianoform scores, which are among the most challenging in graphical terms. The evaluation was designed around two scenarios: a controlled one using synthetic data to preliminarily assess the performance of the model and study the influence of hyperparameters, and a real-world scenario using two scanned music score datasets, where we \review{benchmarked the performance of the system with that of OMR tools as well as layout analysis pipelines}. Results show that our approach not only successfully transcribes pianoform scores but also outperforms \review{state-of-the-art OMR when fine-tuned and shows competitive results with commercial systems when trained exclusively with synthetic data.} Consequently, the model establishes a new state-of-the-art approach for full-page polyphonic music transcription and stands as the first true end-to-end solution for OMR.

This work opens several avenues for improvement and future research. While the model establishes a new benchmark for OMR performance, there are opportunities to refine the curriculum learning pipeline. \review{First, by improving the synthetic generator to provide better and more varied images, as well as providing a richer backbone datasets, given the great limitations \GrandStaff{} shows, especially for the zero-shot scenarios}. \review{In this matter, the integration of music-notation learning would improve the correlation of the language model correlation with musical elements.} Additionally, like all Transformer-based architectures, the model is data-demanding. Future research could focus on optimization strategies and Self-Supervised Learning (SSL) to improve performance and reduce reliance on large labeled datasets.

\begin{appendices}

\section{\review{OMR tools transcription pipeline}}
\label{app:pipeline}
In Section \ref{subsec:commercial_presentation}, we introduced \review{several OMR tools to set a reference with the \FPSMTNXT{} in the real scores scenario}. In this appendix, we detail the pipeline followed to obtain the \krn{} scores for evaluation. This pipeline, depicted in Fig. \ref{fig:photoscore_pipeline}, consists of three main steps:

\begin{figure*}[h]
    \centering
    \includegraphics[width=\textwidth]{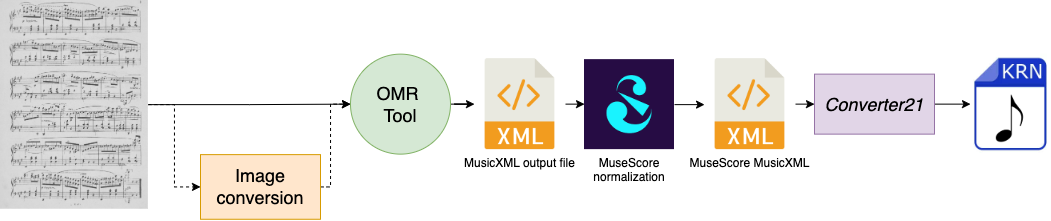}
    \caption{\review{Transcription pipeline for teh OMR tools in the real scores evaluation scenario.}}
    \label{fig:photoscore_pipeline}
\end{figure*}

\begin{enumerate}
    \item \textbf{Image conversion}: \review{The target datasets are provided in JPG and PNG formats, which are compatible with most of the OMR tools. However, the PhotoScore tool only accepts PDF and bitmap files. To ensure a fair comparison, we converted---for this case---all test set images into the TIFF format to make them compatible with the software.}.
    \item \review{\textbf{Transcription}: The converted images are processed using the corresponding tool, which automatically recognizes and transcribes the images. Once the transcription process is complete, results are exported to MusicXML~\citep{Good:XML:2001}, which is the common format for all tools}.
    \item \textbf{Conversion to MuseScore MusicXML}: One of the main challenges of using MusicXML is its flexible syntax for representing music scores, which complicates the eventual conversion to \krn{}. Following the approach of~\cite{Mayer:ICDAR:2024}, we convert the MusicXML format into MuseScore \footnote{\url{https://musescore.org/es}} MusicXML format, a widely adopted standard used by the community and modern music composition and analysis tools. We open the score in MuseScore and automatically convert it into this specific MusicXML format.
    \item \textbf{Conversion to Humdrum Kern}: Finally, we convert the MuseScore MusicXML to \krn{} using the \textit{converter21} library, which, to our knowledge, is the best-performing tool for conversions between music encoding formats.
\end{enumerate}

\review{Note that this pipeline may crash or produce unreadable transcriptions in any of the tools, especially with PhotoScore and Audiveris. If a sample fails to produce a \krn{} file, a 100\% \ac{SER} is given, as the user would have to manually write down the complete transcription of the document.}

\end{appendices}

\bmhead{Acknowledgements}
The first author is supported by grants ACIF/2021/356 and CIBEFP/2022/19 from the ``Programa I+D+i de la Generalitat Valenciana''.
Second author is supported by grant CISEJI/2023/9 from ``Programa para el apoyo a personas investigadoras con talento (Plan GenT) de la Generalitat Valenciana''. Special thanks to Joan Cerveto Serrano and Eliseo Fuentes Martínez for their for their invaluable collaboration in preparing the Polish Digital Scores and Mozarteum datasets.

\section*{Declarations}
\begin{itemize}
    \item \textbf{Funding}: The first author is supported by grants ACIF/2021/356 and CIBEFP/2022/19 from the ``Programa I+D+i de la Generalitat Valenciana''. Second author is supported by grant CISEJI/2023/9 from ``Programa para el apoyo a personas investigadoras con talento (Plan GenT) de la Generalitat Valenciana''.
    \item \textbf{Conflict of interest / Competing interests}: The authors have no competing interests to declare that are relevant to the content of this article.
    \item \textbf{Ethics approval and consent to participate}: Not applicable.
    \item \textbf{Consent for publication}: Not applicable.
    \item \textbf{Data availability}: The \GrandStaff{} and the Full-page \GrandStaff{} datasets are publicly available at the HuggingFace datasets hub (\url{https://huggingface.co/collections/antoniorv6/sheet-music-transformer-datasets-66defa88d50145aa1a518822}). The Mozarteum and the Polish digital scores datasets are available upon formal request to the first author, although they will be publicly released through the HuggingFace hub platform.
    \item \textbf{Materials availability}: Not applicable
    \item \textbf{Code availability}: The source code of the model is open-source. It is available at \url{https://github.com/antoniorv6/SMT-plusplus.git} while the model and the weights are published in the HuggingFace Transformers platform \url{https://huggingface.co/antoniorv6}.
    \item \textbf{Author contribution}: All authors equally contributed to the production of this work.
    
\end{itemize}

%Acknowledgements are not compulsory. Where included they should be brief. Grant or contribution numbers may be acknowledged.

%Please refer to Journal-level guidance for any specific requirements.

%\section*{Declarations}

%Some journals require declarations to be submitted in a standardised format. Please \cmark the Instructions for Authors of the journal to which you are submitting to see if you need to complete this section. If yes, your manuscript must contain the following sections under the heading `Declarations':

%\begin{itemize}
%\item Conflict of interest/Competing interests (\cmark journal-specific guidelines for which heading to use)
%\item Ethics approval and consent to participate
%\item Consent for publication
%\item Data availability 
%\end{itemize}

%\noindent
%If any of the sections are not relevant to your manuscript, please include the heading and write `Not applicable' for that section. 

\bibliography{sn-bibliography}% common bib file

\end{document}